\documentclass[10pt, a4paper]{article}

\usepackage{lrec-coling2024}

\usepackage{graphicx}
\usepackage{amsfonts}
\usepackage{amsmath}
\usepackage{bigdelim}
\usepackage{diagbox}
\usepackage{amsthm}
\usepackage{makecell}
\usepackage{mathtools}
\usepackage{booktabs}
\usepackage{footmisc}
\usepackage{xcolor}
\usepackage{tabularx}
\usepackage[shortlabels]{enumitem}
\graphicspath{ {figs/} }

\newcommand{\quash}[1]{}  
\newcommand{\cready}[1]{} 
\newcommand{\gpt}{\textsc{GPT-2}}
\newcommand{\bert}{\textsc{BERT}}

\newcommand{\mask}{\texttt{[MASK]}}

\newcommand{\mat}{\texttt{mat}}
\newcommand{\id}{\texttt{id}}
\newcommand{\matl}{\texttt{mat}_{\ell \rightarrow \ell'}}
\newcommand{\matll}[2]{\texttt{mat}_{{#1} \rightarrow {#2}}}
\newcommand{\bll}[2]{\texttt{b}_{{#1} \rightarrow {#2}}}
\newcommand{\matattn}{\texttt{mat\_attn}}
\newcommand{\matattnl}{\texttt{mat\_attn}_{\ell \rightarrow \ell'}}
\newcommand{\matff}{\texttt{mat\_ffn}}
\newcommand{\matffl}{\texttt{mat\_ffn}_{\ell \rightarrow \ell'}}
\newcommand{\matln}{\texttt{mat\_ln1\_ln2}}
\newcommand{\matlnl}{\texttt{mat\_ln1\_ln2}_{\ell \rightarrow \ell'}}
\newcommand{\idl}{\texttt{id}_{\ell \rightarrow \ell'}}
\newcommand{\matlL}{\texttt{mat}_{\ell \rightarrow L}}
\newcommand{\matattnlL}{\texttt{mat\_attn}_{\ell \rightarrow L}}
\newcommand{\matfflL}{\texttt{mat\_ffn}_{\ell \rightarrow L}}
\newcommand{\matlnlL}{\texttt{mat\_ln1\_ln2}_{\ell \rightarrow L}}
\newcommand{\idlL}{\texttt{id}_{\ell \rightarrow L}}

\definecolor{light_blue}{HTML}{cfdfff}
\usepackage[most]{tcolorbox}
\tcbset{on line, 
        boxsep=1pt, left=0pt,right=0pt,top=0pt,bottom=0pt,
        colframe=white,colback=light_blue,  
        highlight math style={enhanced}
        }

\title{Jump to Conclusions: Short-Cutting Transformers\\ with Linear Transformations}

\name{Alexander Yom Din$^1$, ~ Taelin Karidi$^1$, ~ Leshem Choshen$^1$, ~ Mor Geva$^2$}

\address{$^1$Hebrew University of Jerusalem ~~~ $^2$Tel Aviv University \\
         \{alexander.yomdin, taelin.karidi, leshem.choshen\}@mail.huji.ac.il, morgeva@tauex.tau.ac.il\\}

\abstract{
Transformer-based language models create hidden representations of their inputs at every layer, but only use final-layer representations for prediction. This obscures the internal decision-making process of the model and the utility of its intermediate representations. One way to elucidate this is to cast the hidden representations as final representations, bypassing the transformer computation in-between.
In this work, we suggest a simple method for such casting, using linear transformations. This approximation far exceeds the prevailing practice of inspecting hidden representations from all layers, in the space of the final layer. 
Moreover, in the context of language modeling, our method produces more accurate predictions from hidden layers, across various model scales, architectures, and data distributions. This allows ``peeking'' into intermediate representations, showing that \gpt{} and \bert{} often predict the final output already in early layers.
We then demonstrate the practicality of our method to recent early exit strategies, showing that when aiming, for example, at retention of 95\% accuracy, our approach saves additional 7.9\% layers for \gpt{} and 5.4\% layers for \bert{}.
Last, we extend our method to linearly approximate sub-modules, finding that attention is most tolerant to this change. Our code and learned mappings are publicly available at \url{https://github.com/sashayd/mat}.
\\ \newline \Keywords{interpretability, language models, efficiency, logitlens, linear lense, linear, early exit, shortcut, layer jump}
}

\begin{document}

\maketitleabstract

\section{Introduction}
\label{sec:introduction}

Transformer-based language models (LMs) process an input sequence of tokens by first representing it as a sequence of vectors and then repeatedly transforming it through a fixed number of attention and feed-forward network (FFN) layers 
\cite{NIPS2017_3f5ee243}. While each transformation creates new representations, only the final representations are used to obtain model predictions. Correspondingly, LM loss minimization directly optimizes the final representations, while hidden representations are only optimized implicitly, thus making their interpretation and usefulness more obscure.

\begin{figure}
\setlength{\belowcaptionskip}{-10pt}
\centering
\includegraphics[scale=0.5]{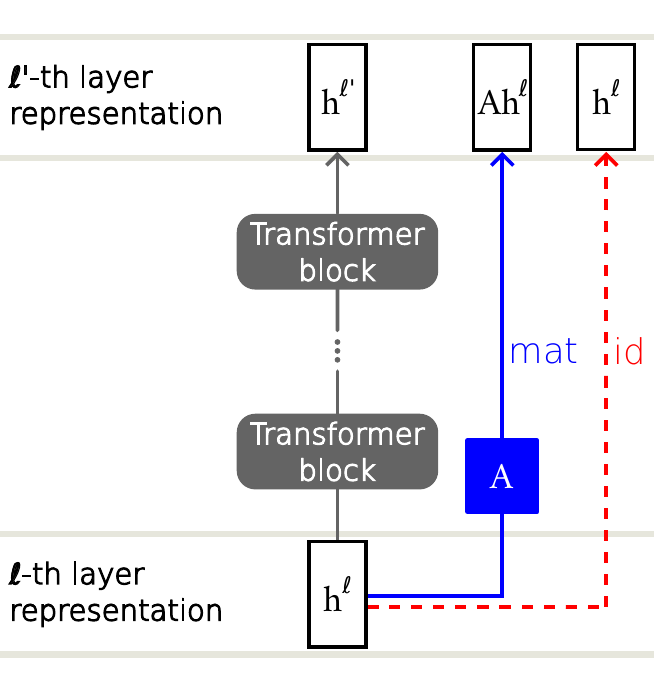}
\caption{An illustration of our approach to enhance interpretability and utilization of hidden representations.  
We use linear mappings $A = A_{\ell',\ell}$ to short-cut transformer inference in-between layers ($\matl{}$), instead of the prevalent baseline of propagating the hidden representation as-is to the further layer ($\idl{}$).}
\label{fig:main}
\end{figure}

However, utilizing hidden representations is highly desirable; a successful interpretation of them can shed light on the ``decision-making process'' in the course of transformer inference \cite{tenney-etal-2019-bert, voita-etal-2019-bottom, slobodkin-etal-2021-mediators,geva-etal-2022-transformer}, and obtaining predictions from them can substantially reduce computational cost \cite{schwartz-etal-2020-right,xu2021survey}.

Previous attempts to harness hidden representations viewed the hidden representations of an input token as a sequence of approximations of its final representation \cite{Elhage, geva-etal-2022-transformer}. This view is motivated by the additive updates induced via the residual connections \cite{He} around each layer in the network. Indeed, previous works \cite{geva-etal-2021-transformer, geva-etal-2022-lm, ram2022token, alammar-2021-ecco} followed a simplifying assumption that representations \textit{at any layer} can be transformed into a distribution over the output vocabulary via projection to the output embeddings.
While this approach has proven to be surprisingly effective for interpretability \cite{geva-etal-2022-lm, dar2022analyzing} and computation efficiency \cite{schuster2022confident, xin-etal-2020-deebert, schwartz-etal-2020-right}, it oversimplifies the model's computation and assumes that all the layers in the network operate in the same space.

A natural question that arises is whether there is a more accurate way to cast hidden representations into final representation substitutes than interpreting them as they are. 
In this work, we tackle this question by learning linear transformations across layers in the network (illustrated in Fig.~\ref{fig:main}). For any two layers $\ell < \ell'$, we fit a linear regression to transform hidden representations from layer $\ell$ to layer $\ell'$. 
We show that this method, denoted as \mat{}, produces substantially more accurate approximations than the above-discussed identity mapping, dubbed \id{}, applied in previous works (\S\ref{sec:layer_jump}).
As \mat{} is a non-contextual mapping that operates on single hidden representations, 
this suggests that there is more linearity to transformer inference than could be estimated by the \id{} mapping.

Next, we test if these gains in approximating future representations also translate to better prediction estimations (\S\ref{sec:prediction}).
To this end, we
measure 
how often language modeling predictions from final representation substitutes produced by \mat{}, and by alternations between \mat{} and regular inference, agree with those of actual final representations. 
Through experiments with two data sources and various scales of \gpt{} \cite{radford2019language} and \bert{} \cite{devlin-etal-2019-bert}, we observe large accuracy gains ($15\%$-$40\%$ at most layers) in prediction estimation by \mat{} over naive projections (\id{}).
Moreover, we show that our mappings generalize well across different data distributions (\S\ref{sec:robustness}).

We leverage these findings for enhancing model efficiency and demonstrate our method's utility in the setting of early exiting -- a strategy for dynamically deciding at which layer to stop the inference pass and use that layer's representation for prediction.
While previous works have utilized these hidden representations intact (i.e. using \id{}), we transform them using \mat{}, showing that our method performs better than the baseline in this setting as well (\S\ref{sec:applications}), allowing for the saving of additional $7.9\%$ (resp. $5.4\%$) of the layers for \gpt{} (resp. \bert{}) when aiming at $95\%$ accuracy.

Last, we analyze how well the different sub-modules of transformer computation -- attention, FFN, and layer normalization -- can be estimated linearly (\S\ref{sec:submodules}), by applying the same methodology of linear mappings.
We find that linearly approximating attention, the only sub-module that has contextual processing, results in the least reduction of precision. This hints at an interesting possibility of compute time reduction, as non-contextual inference is parallelizable.

To conclude, we propose a method for casting hidden representations across transformer layers, that is light to train, cheap to infer, and provides more accurate and robust representation approximations than the commonly-used
baseline of identical propagation.
Beyond interpretability, our method holds potential for enhancing efficiency.
\section{Background and Notation}
\label{sec:background}

The input to a transformer-based LM \cite{NIPS2017_3f5ee243} is a sequence of tokens $t_1, ..., t_n$ from a vocabulary $\mathcal{V}$ of size $|\mathcal{V}| = d_v$. The tokens are first represented as vectors using an embedding matrix $E \in \mathbb{R}^{d_h \times d_v}$, where $d_h$ is the hidden dimension of the model, to create the initial \emph{hidden representations} $$ H^0 = ( h_1^0, \ldots, h_n^0 ) \in \mathbb{R}^{d_h \times n}.$$ 
These representations are then repeatedly transformed through $L$ transformer blocks, where each block outputs hidden representations that are the inputs to the next block:
$$ \forall \ell \in [1, L]: \;\; \texttt{b}^\ell (H^{\ell-1}) = H^{\ell} $$ 
where $$H^\ell = (h^{\ell}_1, \ldots, h^{\ell}_n) \in \mathbb{R}^{d_h \times n}.$$

The $\ell$-th transformer block is constructed as a composition of two layers: $$ \texttt{b}_\ell = \texttt{b}^{\texttt{ffn}}_\ell \circ \texttt{b}^{\texttt{attn}}_\ell,$$ where $\texttt{b}^{\texttt{attn}}_\ell$  (resp. $\texttt{b}^{\texttt{ffn}}_\ell$) is a multi-head self-attention (MHSA) layer (resp. FFN layer) enclosed by a residual connection, and potentially interjected with layer normalization \cite{ba2016layer}.
The \emph{final representations},
$$ H^L = ( h_{1}^{L}, \ldots, h_{n}^{L}), $$ are
the transformer stack's output, used to form various predictions. In this work, we
investigate whether and how hidden representations from \textit{earlier layers} can be utilized for this purpose instead.
\section{Linear Shortcut Across Blocks}
\label{sec:layer_jump}

To use a hidden representation from layer $\ell<L$ as a final representation, we propose to cast it using linear regression, while skipping the computation in-between these layers. More generally, this approach can be applied to cast any $\ell$-th hidden representation to any subsequent layer $\ell'>\ell$.

\subsection{Method}
\label{subsec:methodology_linear_shortcut}

Given a source layer $\ell$ and a target layer $\ell'$ such that $0 \leq \ell < \ell' \leq L$, our goal is to learn a mapping
from hidden representations at layer $\ell$ to those at layer $\ell'$. To this end, we first collect a set of corresponding hidden representation pairs $(h^\ell, h^{\ell'})$. Concretely, we run a set $\mathcal{T}$ of input sequences through the model, and for each input $s$, we extract the hidden representations $h_{i_s}^{\ell}, h_{i_s}^{\ell'}$, where $i_s$ is a random position in $s$.
Next, we learn a matrix $A_{\ell', \ell} \in \mathbb{R}^{d_h \times d_h}$ by fitting linear regression over $\mathcal{T}$, i.e., $A_{\ell', \ell}$ is a numerical minimizer for:
$$ A \mapsto \sum_{s \in \mathcal{T}} || A \cdot h_{i_s}^\ell - h_{i_s}^{\ell'} ||^2,$$ 
and define the mapping of a representation $h$ from layer $\ell$ to layer $\ell'$ as:
\begin{equation}
\label{eq:linear_jump}
    \matl{} (h) \coloneqq A_{\ell', \ell} \cdot h.
\end{equation}

\subsection{Baseline}
\label{subsec:baseline}

We evaluate 
the prevalent approach of ``reading'' hidden representations directly, without any transformation. 
Namely, the propagation of a hidden representation from layer $\ell$ to layer $\ell'$ is given by the identity function, dubbed \id{}:

$$ \idl{} (h) \coloneqq h.$$

This baseline 
assumes that representations at different layers operate in the same linear space.

\subsection{Quality of Fit}
\label{subsec:experiments_r2}

We first evaluate our method by measuring how well the learned linear mappings approximate the representations at the target layer. To this end, we calculate the (coordinate-averaged) $r^2$-score of our mapping's outputs with respect to the representations obtained from a full inference pass, and compare to the same for the \id{} baseline.

\paragraph{Models.}

We use \gpt{} \cite{radford2019language}, a decoder-only auto-regressive LM, with $L = 48$, $d_h = 1600$, and \bert{} \cite{devlin-etal-2019-bert}, an encoder-only model trained with masked language modeling, with $L=24$, $d_h=1024$.

\paragraph{Data.}
We sample random sentences from Wikipedia,
collecting 9,000 (resp. 3,000) sentences for the training set $\mathcal{T}$ (resp. validation set $\mathcal{V}$).\footnote{We use sentences rather than full documents to simplify the analysis.}
For each example $s$, we select a random position $i_s$ and extract the hidden representations $h_{i_s}^{\ell}$ at that position from all the layers.
For \bert{}, we first replace the input token at position $i_s$ with a \mask{} token, as our motivation is interpreting predictions, which are obtained via masked tokens in \bert{} (see \S\ref{subsec:BERT}).
Thus, in this case, the hidden representations we consider
are of \mask{} tokens only.

\begin{figure}[t]
\includegraphics[scale=0.2]{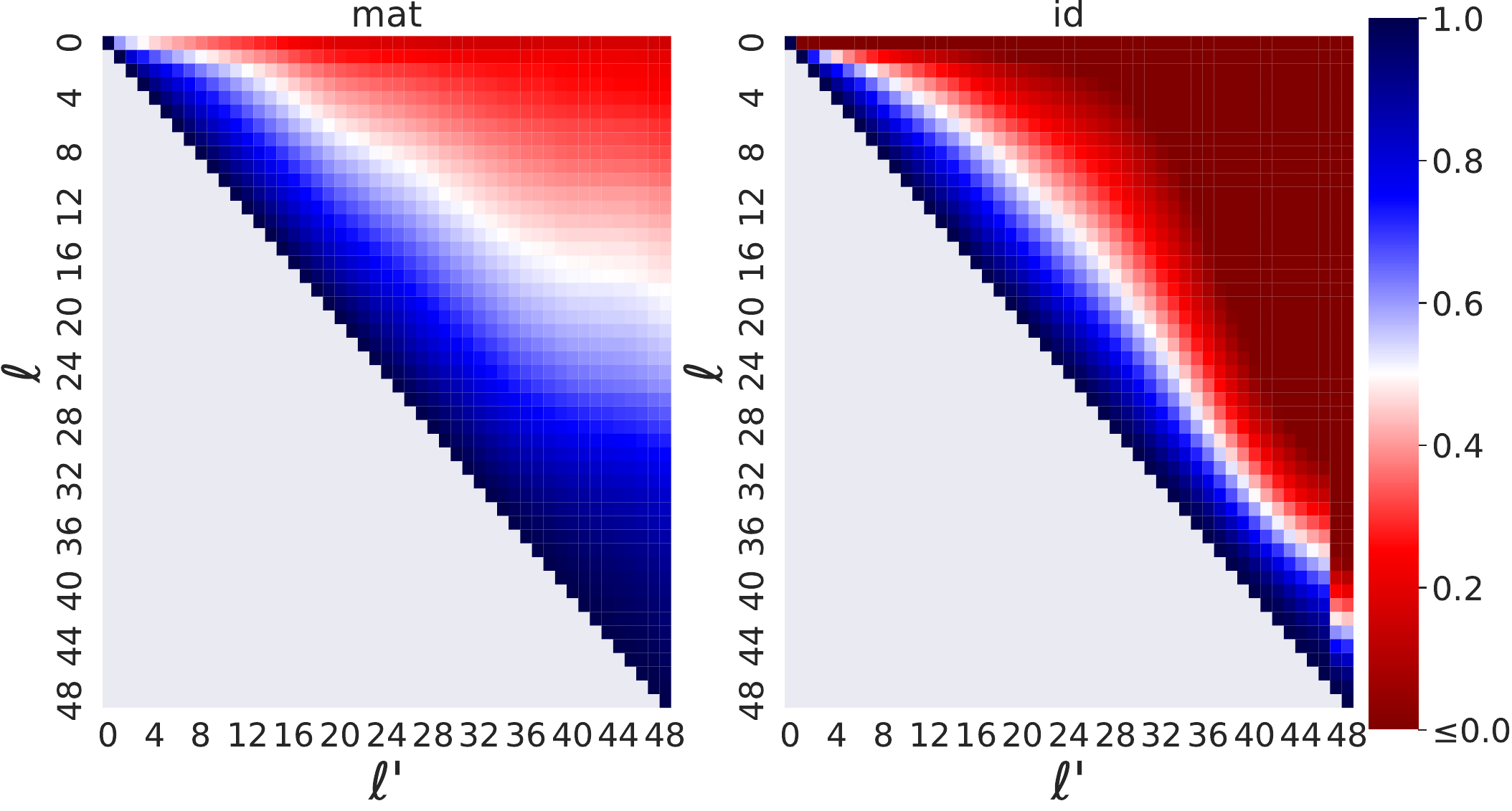}
\caption{The coordinate-averaged $r^2$-score of $\matl{}$ (left) and $\idl{}$ (right) (\gpt{}).}
\label{fig:r2_scores}
\end{figure}

\begin{figure}[t]
\setlength{\belowcaptionskip}{-10pt}
\includegraphics[scale=0.2]{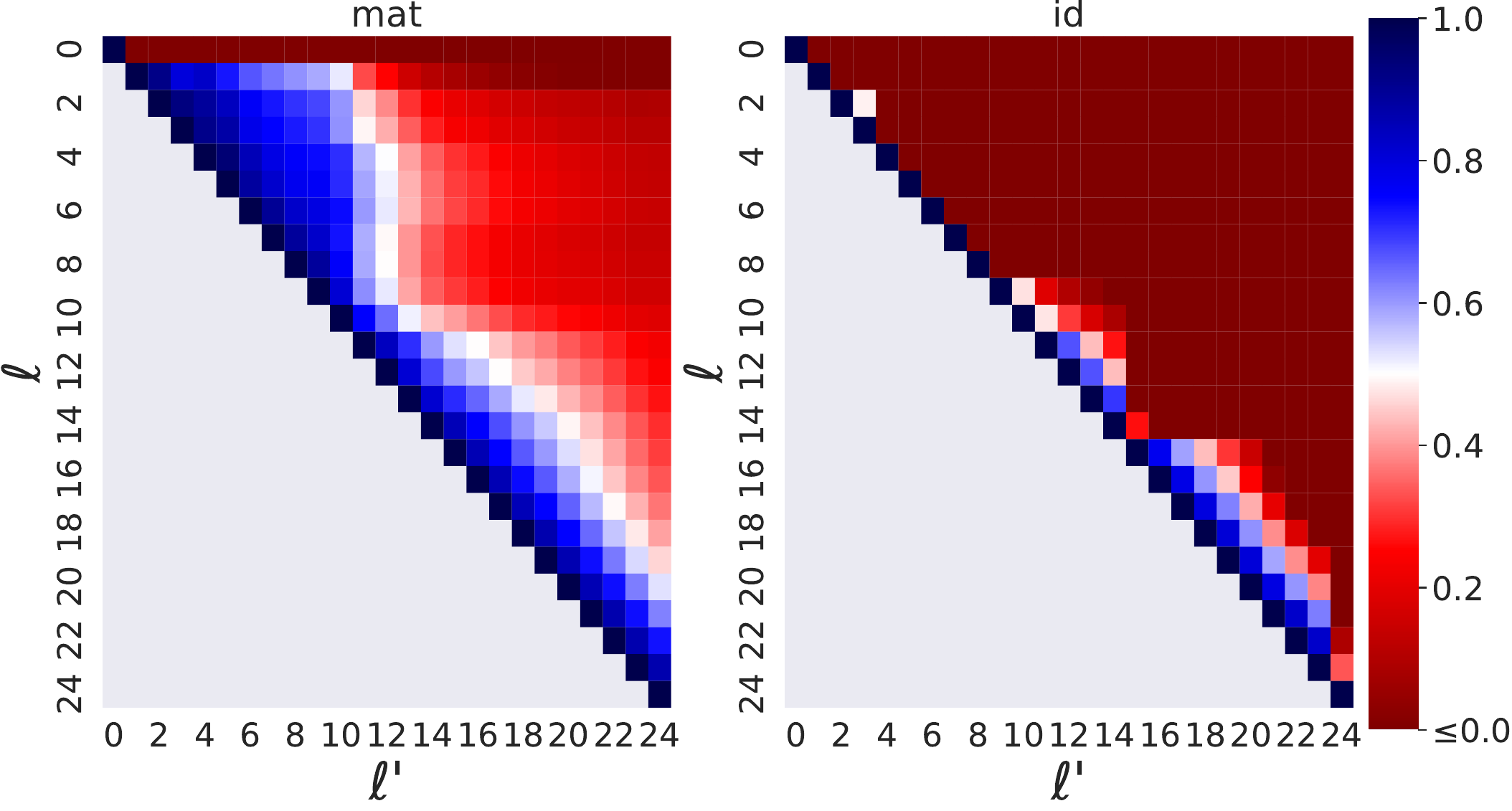}
\caption{The coordinate-averaged $r^2$-score of $\matl{}$ (left) and $\idl{}$ (right) (\bert{}).}
\label{fig:bertmask_r2_scores}
\end{figure}

\paragraph{Evaluation.}
For every pair of layers $\ell, \ell'$, such that $0 \leq \ell < \ell' \leq L$, we use the training set $\mathcal{T}$ to fit linear regression as described in \S\ref{subsec:methodology_linear_shortcut}, and obtain a mapping $\matl{}$. 
Next, we evaluate the quality of $\matl{}$ as well as of $\idl{}$ using the $r^2$-coefficient, uniformly averaged over all coordinates. Concretely, we compute the $r^2$-coefficient of each of the predicted representations $\matl{} (h_{i_s}^{\ell})$ and $\idl{} (h_{i_s}^{\ell})$ versus the true representations $h_{i_s}^{\ell'}$
over all $s \in \mathcal{V}$.

\paragraph{Results.}
Results for \gpt{} and \bert{} are presented in Figs.~\ref{fig:r2_scores} and~\ref{fig:bertmask_r2_scores}, respectively.
In both models, \mat{} consistently yields better approximations than \id{}, as it obtains higher $r^2$-scores (in blue) across the network. 
This gap between \mat{} and \id{} is especially evident in \bert{}, where \id{} completely fails to map the representations between most layers, suggesting that hidden representations are modified  substantially by every transformer block.
Overall, this highlights the shortcoming of existing practices to inspect representations in the same linear space, and the gains from using our method to approximate future layers.
\section{Linear Shortcut for Language Modeling}
\label{sec:prediction}

We saw that our method approximates future hidden representations substantially better than a naive propagation. 
In this section, we show that this improvement also translates to better predictive abilities from earlier layers. Concretely, we use our method to estimate the final prediction from intermediate representations, in the context of two fundamental LM tasks; next token prediction and masked token prediction.

\paragraph{Evaluation Metrics.}
Let $h, h' \in \mathbb{R}^{d_h}$ be a final representation and its  substitute obtained by some mapping, and denote by $\delta (h), \delta (h') \in \mathbb{R}^{d_v}$ their corresponding output probability distributions
(see details below).
We measure the prediction quality of $h'$ with respect to $h$ using two metrics:
\begin{itemize}
[leftmargin=*,topsep=2pt,parsep=1pt]
    \item \textbf{Precision@$k$} ($\uparrow$ is better): This checks whether the token with the highest probability according to $\delta(h')$ appears in the top-$k$ tokens according to $\delta(h)$. Namely, we sort $\delta(h)$ and assign a score of $1$ if $\arg\max(\delta(h'))$ appears in the top-$k$ tokens by $\delta(h)$, and $0$ otherwise.
    
    \item \textbf{Surprisal} ($\downarrow$ is better): We measure the negative log likelihood according to $\delta(h)$, of the highest-probability token according to $\delta(h')$. Intuitively, low values mean that the model sees the substitute result as probable and hence not surprising.
\end{itemize}

\noindent We report the average Precision@$k$ and Surprisal over the validation set $\mathcal{V}$.

\subsection{Next Token Prediction}
\label{subsec:next_token_prediction_task}

Auto-regressive LMs output for every position a probability distribution over the vocabulary for the next token. Specifically, the output distribution for every position $i$ is given by $\delta (h_i^L)$, where
\begin{equation}\label{eq:output_distribution}
    \delta (h) = \texttt{softmax} ( E^\top \cdot h) \in \mathbb{R}^{d_v}.
\end{equation}
For some LMs, including \gpt{}, a layer normalization $\texttt{ln\_f}$ is applied to the final layer representation before this conversion (i.e., computing $\delta (\texttt{ln\_f}(h))$ rather than $\delta (h)$).

Recall that our goal is to measure how well this distribution can be estimated from intermediate representations, i.e. estimating $\delta (h_i^L)$ from
$h_i^{\ell}$
where $\ell<L$. 
Thus, we first run the validation set examples through the model, while extracting for each example $s$ and every layer the hidden representation at a random position $i_s$. Next, we apply our mappings $\matlL{}$ and $\idlL{}$ to cast the hidden representations of every layer $\ell$ to final layer substitutes (see \S\ref{sec:layer_jump}). 
Last, we convert every final-layer substitute to an output distribution (Eq.~\ref{eq:output_distribution}) and compute for each layer the average Precision@$k$ (for $k=1,5,10$) and Surprisal scores with respect to the final output distribution, over the validation set.

\begin{figure}[t]
\setlength{\belowcaptionskip}{-10pt}
\centering
\includegraphics[width=\columnwidth]{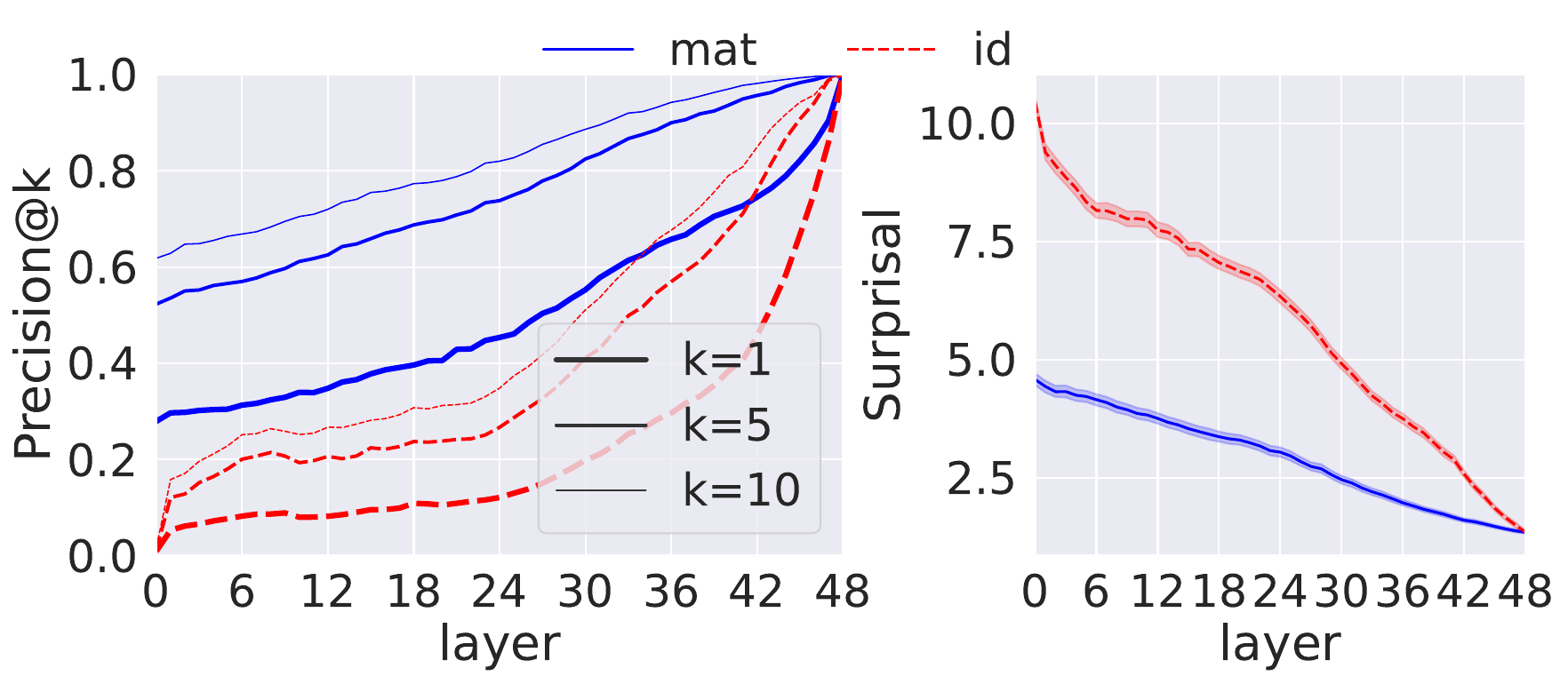}
\caption{Precision@$k$ ($k = 1,5, 10$) and Surprisal for $\matlL{}$ and $\idlL{}$ (\gpt{} next token prediction task). 95\% confidence intervals are shown for Surprisal.}
\label{fig:presurp}
\end{figure}

\quash{
\begin{figure}[t]
\centering
\includegraphics[scale=0.4]{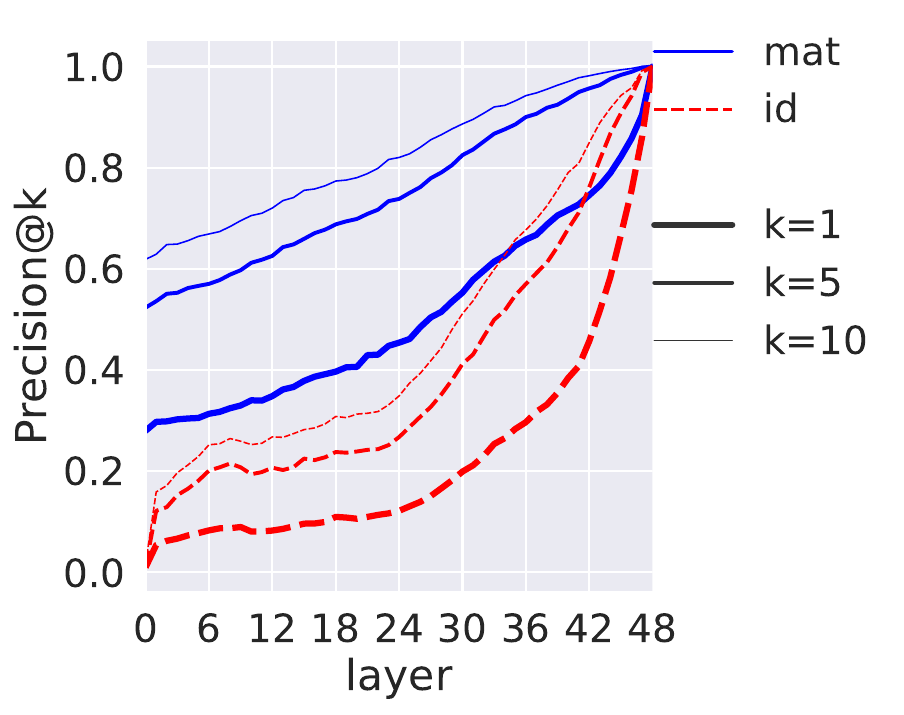}
\caption{Precision@$k$ ($k = 1,5, 10$) for $\matlL{}$ and $\idlL{}$ (\gpt{} next token prediction task).}
\label{fig:pre}
\end{figure}

\begin{figure}[t]
\centering
\includegraphics[scale=0.35]{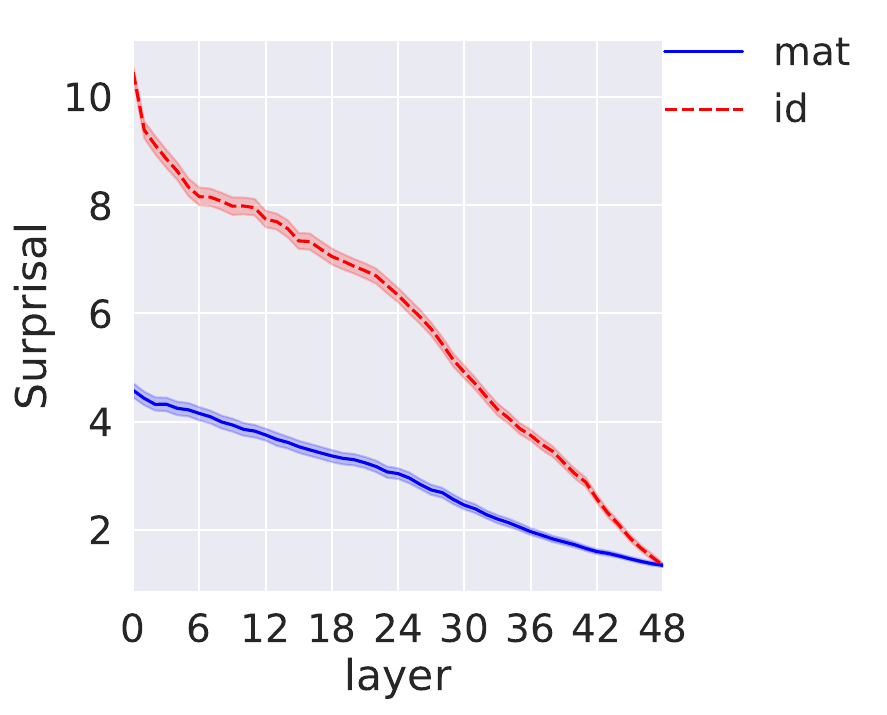}
\caption{Surprisal for $\matlL$ and the baseline $\idlL{}$ (\gpt{} next token prediction task). A 95\% confidence interval surrounds the lines.}
\label{fig:surp}
\end{figure}
}

\paragraph{Results.}
Fig.~\ref{fig:presurp} shows the average Precision@$k$ and Surprisal scores per layer in \gpt{}. Across all layers, \mat{} outperforms \id{} in terms of both scores, often by a large margin (e.g. till layer $44$ the Precision@$1$ achieved by \mat{} is bigger than that of $\id{}$ by more than $20\%$). 
This shows that linear mappings enable not just better estimation of final layer representations, but also of the predictions they induce. Moreover, the relatively high Precision@$k$ scores of \mat{} in early layers ($62\%$-$82\%$ for $k=10$, $52\%$-$74\%$ for $k=5$, and $28\%$-$45\%$ for $k=1$) suggest that early representations often accurately approximate the final prediction. Also, the substantially lower Surprisal scores of \mat{} compared to \id{} imply that our method allows for a more representative reading into the layer-wise prediction-formation of the model than allowed via direct projection to the vocabulary.

\subsection{Masked Token Prediction}
\label{subsec:BERT}

We conduct the same experiment in \S\ref{subsec:next_token_prediction_task} for masked language modeling, where the model predicts a probability distribution for a masked token in the input. 
Unlike next token prediction, where the output distribution is computed from representations of varying input tokens, in masked token prediction the output is always obtained from representations of the same input token (i.e. \texttt{[MASK]}).

For this experiment, we use \bert{}, on top of which we use a pretrained masked language model head $\delta$; given a token sequence $s$, a \mask{} token inside it and its final representation $h$, $\delta (h) \in \mathbb{R}^{d_v}$
 is a probability distribution over tokens giving the model's assessment
 of the likelihood of tokens to be fitting in place of the \mask{} token in $s$.

\begin{figure}[t]
\setlength{\belowcaptionskip}{-10pt}
\centering
\includegraphics[width=\columnwidth]{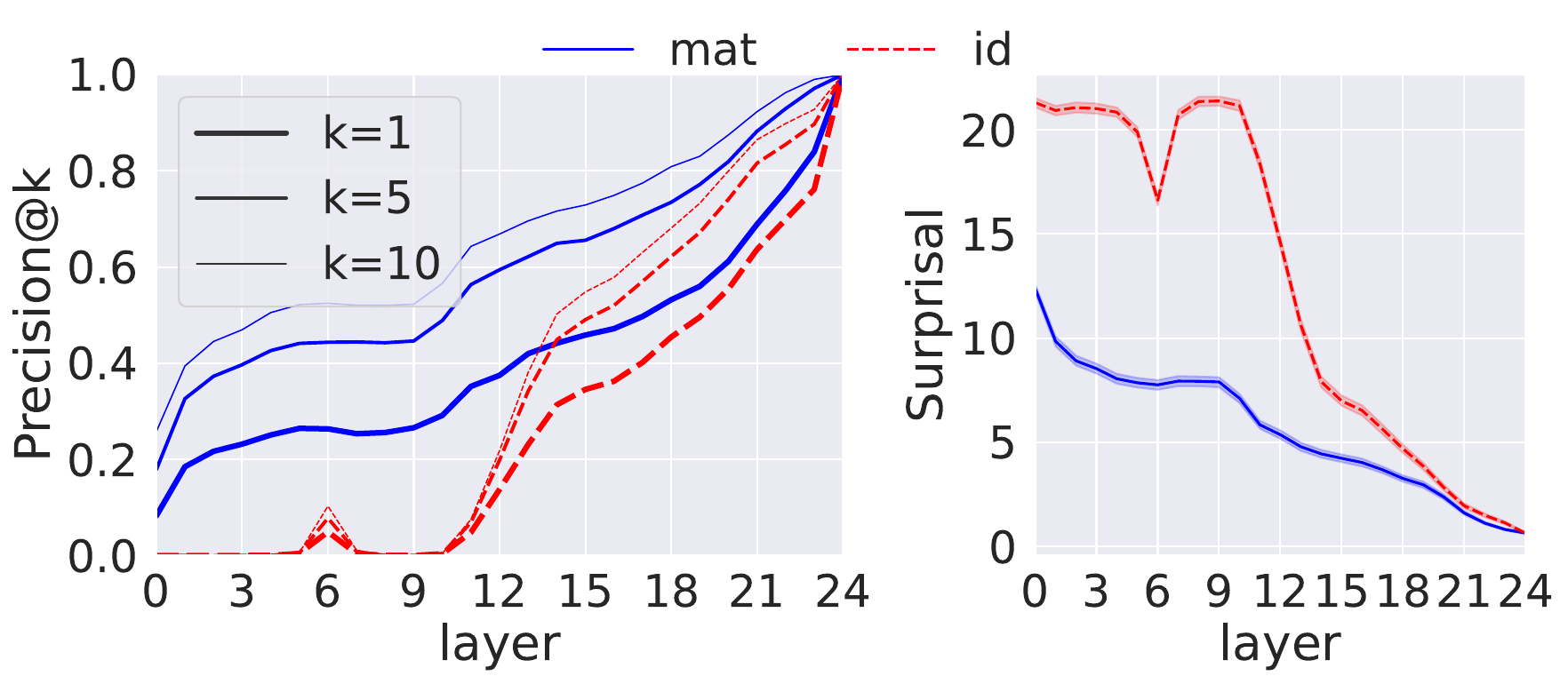}
\caption{Precision@$k$ ($k = 1,5, 10$) and Surprisal for $\matlL{}$ and $\idlL{}$ (\bert{} masked token prediction task). 95\% confidence intervals are shown for Surprisal.}
\label{fig:bertmask_presurp}
\end{figure}

\quash{
\begin{figure}[t]
\centering
\includegraphics[scale=0.4]{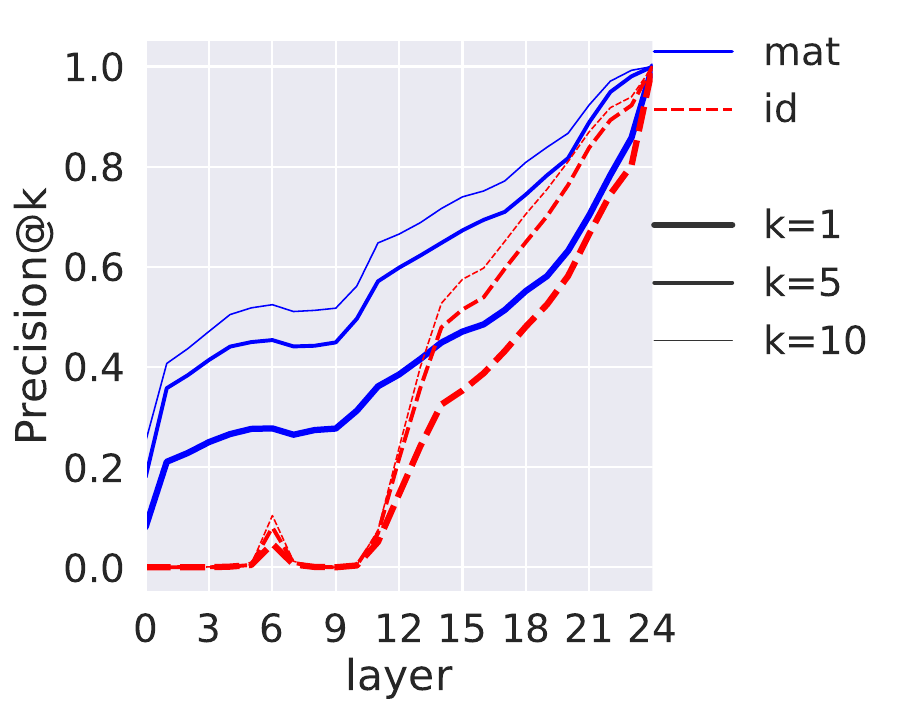}
\caption{Precision@$k$ ($k = 1,5, 10$) for  $\matlL{}$ and the baseline $\idlL{}$ (\bert{} masked token prediction task).}
\label{fig:bertmask_pre}
\end{figure}

\begin{figure}[t]
\centering
\includegraphics[scale=0.35]{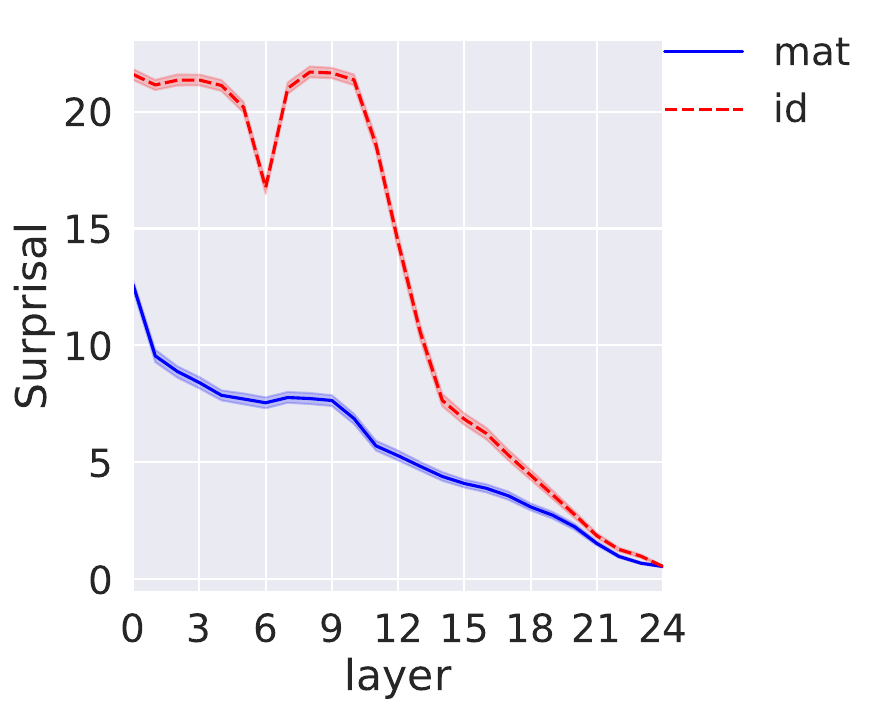}
\caption{Surprisal for $\matlL{}$ and the baseline $\idlL{}$ (\bert{} masked token prediction task). A 95\% confidence interval surrounds the lines.}
\label{fig:bertmask_surp}
\end{figure}
}

\paragraph{Results.}
Fig.~\ref{fig:bertmask_presurp} shows the average Precision@$k$ and Surprisal scores per layer in \bert{}, overall showing trends similar to those observed for next token prediction in \gpt{} (\S\ref{subsec:next_token_prediction_task}). This is despite the differences between the two tasks and the considerable architectural differences between the models.
Notably, the superiority of \mat{} over \id{} in this setting is even more prominent; 
while, in the first ten layers, \mat{}'s precision is between $8\%$-$52\%$  (Fig.~\ref{fig:bertmask_presurp}), \id{}'s precision for all values of $k$ is close to zero, again strongly indicating that our method allows for better reading into early layer hidden representations. 
More generally, \mat{} improves the Precision@$1$ of \id{} by more than $17\%$ at most layers, and unveils that a substantial amount of predictions ($>25\%$ starting from layer $3$) appear already in the very first layers.
Interestingly, the (rough) divide between the first last halves of layers for $\id{}$ in Fig.~\ref{fig:bertmask_presurp} seems to align with the two-hump shape of the blue region for $\mat{}$ in Fig.~\ref{fig:bertmask_r2_scores}.

\begin{table}
\setlength{\belowcaptionskip}{-10pt}
    \footnotesize
    \begin{tabularx}{\columnwidth}
    {p{1.3cm}lp{1.4cm}p{0.7cm}p{1.4cm}}
        $\texttt{id}_{4}$ & $\texttt{mat}_{4}$ & $\texttt{id}_{12}$ & $\texttt{mat}_{12}$ & $\texttt{id}_{24}$ \\ \midrule[1pt]
        \multicolumn{5}{l}{Input: \textit{aldridge had shoulder surgery in \mask{}.}} \\ \midrule
        fellowship & \tcbox{time} & cyclist & \tcbox{2009} & \tcbox{september} \\
        employment & \tcbox{it} & emergencies & \tcbox{2008} & \tcbox{november} \\
        agreement & her & seniors & \tcbox{2010} & \tcbox{december} \\
        \#\#ostal & them & cycling & \tcbox{2006} & \tcbox{august} \\
        \#\#com & work & \tcbox{pennsylvania} & \tcbox{2007} & \tcbox{july} \\ \midrule[1pt]
        \multicolumn{5}{p{7cm}}{Input: \textit{on your next view you will be asked to \mask{} continue reading.}} \\ \midrule
        \#\#com & be & be & be & \tcbox{please} \\
        accreditation & get & undergo & \tcbox{please} & \tcbox{simply} \\ 
        $\copyright$ & go & spartans & help & \tcbox{also} \\ 
        fellowship & \tcbox{help} & seniors & \tcbox{simply} & \tcbox{again} \\ 
        summer & have & * & say & \tcbox{immediately} \\ \bottomrule
    \end{tabularx}
    \caption{Examples of top-$5$ \bert{} masked token predictions at layers $4$, $12$ and $24$, under the mappings $\matlL{}$ (abbreviated $\mat{}_{\ell}$) and $\idlL{}$ (abbreviated $\id{}_{\ell}$). Plausible predictions (according to a human annotator) are marked in \tcbox{blue}. Note that for $\ell=L=24$, predictions of $\mat{}_{\ell}$ and $\id{}_{\ell}$ are the same.}
    \label{tab:manual}
\end{table}

\paragraph{Analysis.}
We manually compare the predictions of our mapping $\matlL{}$ with $\idlL{}$, for the \bert{} model.  Concretely, we select $50$ random sentences from the Leipzig dataset (see \S\ref{subsec:robust_datasets}). Next, for each layer $\ell$, we manually analyze how many of the top-$5$ tokens according to $\matlL{}$ and $\idlL{}$ fit into context. We consider a token to fit into context if it is grammatically plausible within the sentence (see Tab.~\ref{tab:manual} for examples).
In the resulting $1,250$ instances (i.e. $50$ sentences $\times$ $25$ representations), we observe a substantially higher plausibility rate of $85.4\%$ for \mat{} compared to $52.8\%$ for \id{}. In fact, only in less than $4.3\%$ of the instances, there are more plausible tokens among the top-$5$ tokens according to \id{} than according to \mat{}, further supporting the Surprisal results above.

\subsection{Alternation Schemes}
\label{subsec:alternating}
Thus far, we considered direct mappings to the last layer. We now check if mappings between intermediate layers can 
improve prediction estimations further. 
To this end, we obtain final-representation substitutes by alternating between transformer-inference and linear mappings.
For $\ell < \ell'$, let us abbreviate
$$ \bll{\ell}{\ell'} := \texttt{b}^{\ell'} \circ \cdots \circ \texttt{b}^{\ell + 2} \circ \texttt{b}^{\ell + 1},$$
i.e. $\bll{\ell}{\ell'}$ is the application of transformer inference from layer $\ell$ to layer $\ell'$.
For a sequence $0 = \ell_0 < \ell_1 < \ldots < \ell_n = L$, we consider either of \begin{equation}\label{eq:alt} \cdots \circ \bll{\ell_2}{\ell_3} \circ \matll{\ell_1}{\ell_2} \circ \bll{\ell_0}{\ell_1},\end{equation}\begin{equation}\label{eq:alt1} \cdots \circ \matll{\ell_2}{\ell_3} \circ \bll{\ell_1}{\ell_2} \circ \matll{\ell_0}{\ell_1}.\end{equation}
In other words, those are inference modes alternating between transformer inference and application of our linear mappings in a prescribed manner.
We then collect two sets of alternation schemes:
\begin{itemize}
[leftmargin=*,topsep=2pt,parsep=1pt]
    \item \textbf{$r^2$-informed (R2)}: We define the $r^2$-score of Eq.~\ref{eq:alt} (resp. Eq.~\ref{eq:alt1}) to be the product of the $r^2$-scores of  $\matll{\ell_i}{\ell_{i+1}}$ (computed in \S\ref{subsec:experiments_r2}), for $i = 1, 3, \ldots$ (resp. $i = 0, 2, \ldots$). For each $\ell$, we consider the scheme that employs $\ell$ transformer blocks with the maximal $r^2$-score.
    
    \item \textbf{Weighted round-robin (WRR)}: For $a,b \ge 1$ such that $a+b$ divides $L$, we consider $(\ell_0, \ell_1, \ldots )$ given by $(0, a, a+b, 2a+b, 2a+2b, \ldots)$ and the two corresponding schemes Eq.~\ref{eq:alt},~\ref{eq:alt1}.
    In other words, here we alternate between performing $a$ transformer blocks and application of our linear mapping across $b$ layers, for some fixed values of $a$ and $b$.
\end{itemize}

\noindent For the experiment, we use \gpt{} with $24$ layers
(see \S\ref{subsec:robust_scale}) and measure Precision@$1$.

\begin{figure}[t]
\setlength{\belowcaptionskip}{-10pt}
\centering
\includegraphics[scale=0.35]{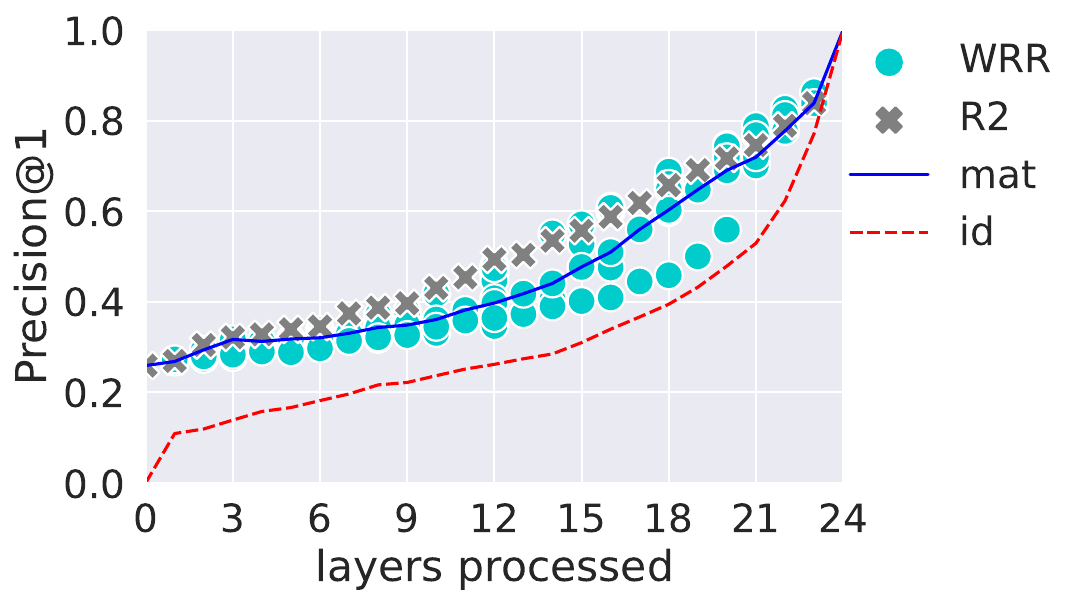}
\caption{Precision@$1$ for various alternation schemes and previous mappings for comparison ($24$-layer \gpt{} next token prediction task).}
\label{fig:alt_24}
\end{figure}

\paragraph{Results.}
Fig.~\ref{fig:alt_24} presents Precision@$1$ for various alternation schemes. We see, first of all, that some alternation schemes provide better precision than the previously considered $\matlL{}$. Second, the best-$r^2$-score tactic for choosing an alternation scheme seems to work well for the first half of layers, but under-achieves (relative to other possible alternation schemes) for the second half. It is, therefore, interesting to try to devise more clever tactics in the future; perhaps, for example, by weighting $r^2$-scores according to layer index.

\section{Method Robustness}
\label{sec:robustness}



\subsection{Robustness Across Model Scales}
\label{subsec:robust_scale}

We repeat our experiments in \S\ref{sec:prediction}, with three additional scales of \gpt{} and one additional scale of \bert{}. Overall, the models are $\texttt{gpt2}$ ($L = 12$, $d_h = 768$), $\texttt{gpt2-medium}$ ($L = 24$, $d_h = 1024$), $\texttt{gpt2-large}$ ($L = 36$, $d_h = 1280$) and $\texttt{gpt2-xl}$ ($L = 48$, $d_h = 1600$), and \texttt{bert-base-uncased} ($L=12$, $d_h=768$) and \texttt{bert-large-uncased} ($L=24$, $d_h=1024$).

\begin{figure}[t]
\centering
\includegraphics[width=\columnwidth]{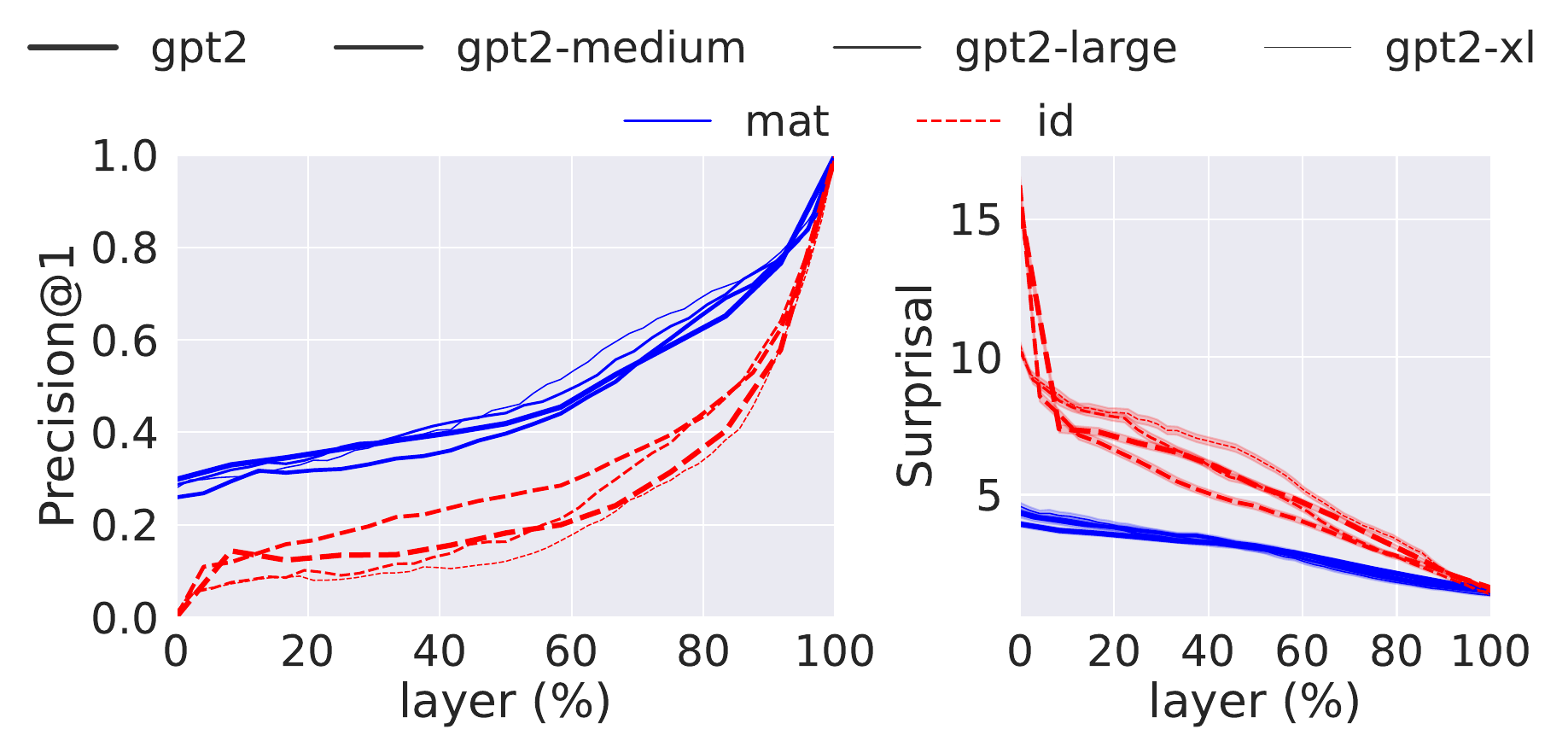}
\caption{Precision@$1$ and Surprisal for $\matlL$ and $\idlL{}$, for next token prediction with \gpt{}. 95\% confidence intervals are shown for Surprisal.}
\label{fig:rob_presurp}
\end{figure}

\begin{figure}[t]
\setlength{\belowcaptionskip}{-10pt}
\centering
\includegraphics[width=\columnwidth]{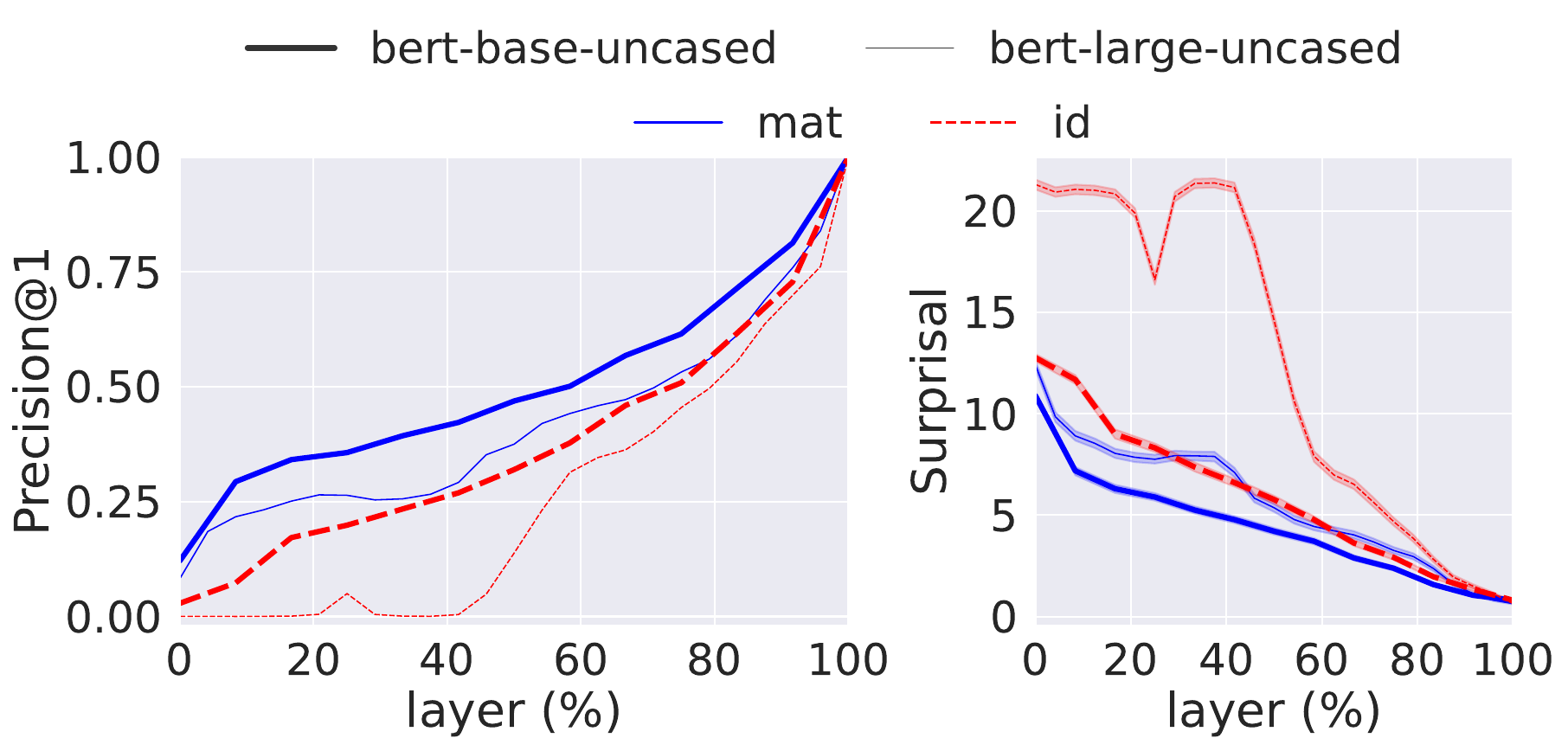}
\caption{Precision@$1$ and Surprisal for $\matlL$ and $\idlL{}$, for masked token prediction with \bert{}. 95\% confidence intervals are shown for Surprisal.}
\label{fig:rob_bert_presurp}
\end{figure}

Fig.~\ref{fig:rob_presurp} (resp. Fig.~\ref{fig:rob_bert_presurp}) depicts the Precision@$1$ and Surprisal scores as functions of the relative depth of the model (i.e. $\ell / L$), for \gpt{} models (resp. \bert{} models).
The plots show the same trends observed in \S\ref{sec:prediction} across various model scales, with \mat{} exhibiting substantially higher predictive abilities from intermediate layers than \id{}.
Interestingly, there is a great overlap between \gpt{} scores of different scales, but not between the scores of \bert{} models.


\quash{
\begin{figure}[t]
\centering
\includegraphics[scale=0.35]{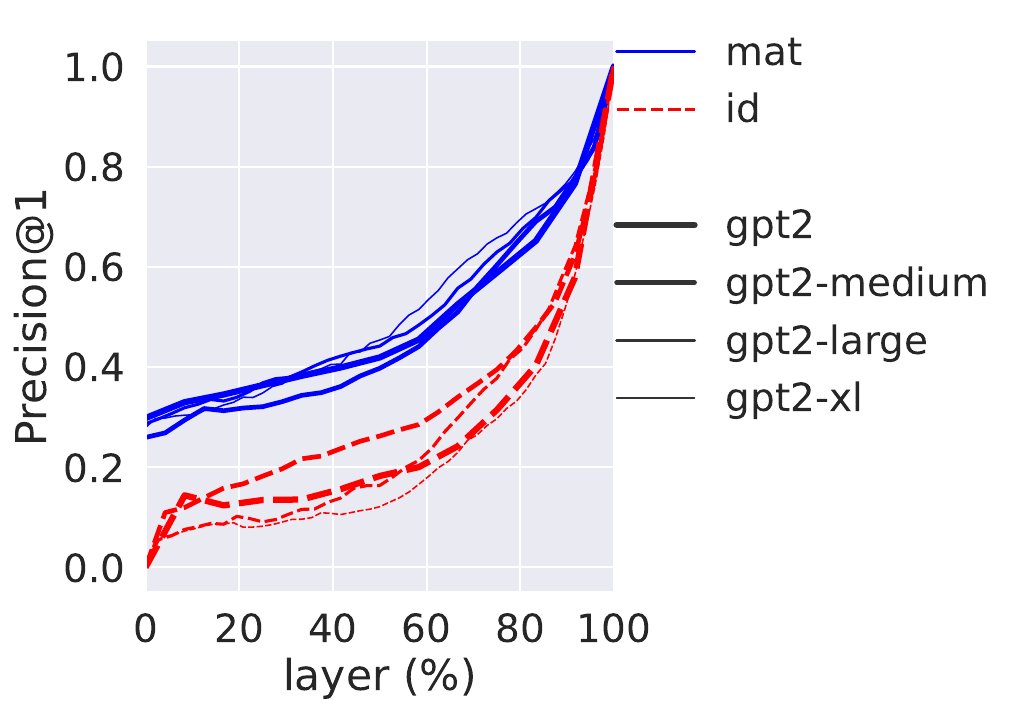}
\caption{Precision@$1$ for $\matlL{}$ and $\idlL{}$ (\gpt{} next token prediction task, four model sizes).}
\label{fig:rob_pre}
\end{figure}

\begin{figure}[t]
\centering
\includegraphics[scale=0.35]{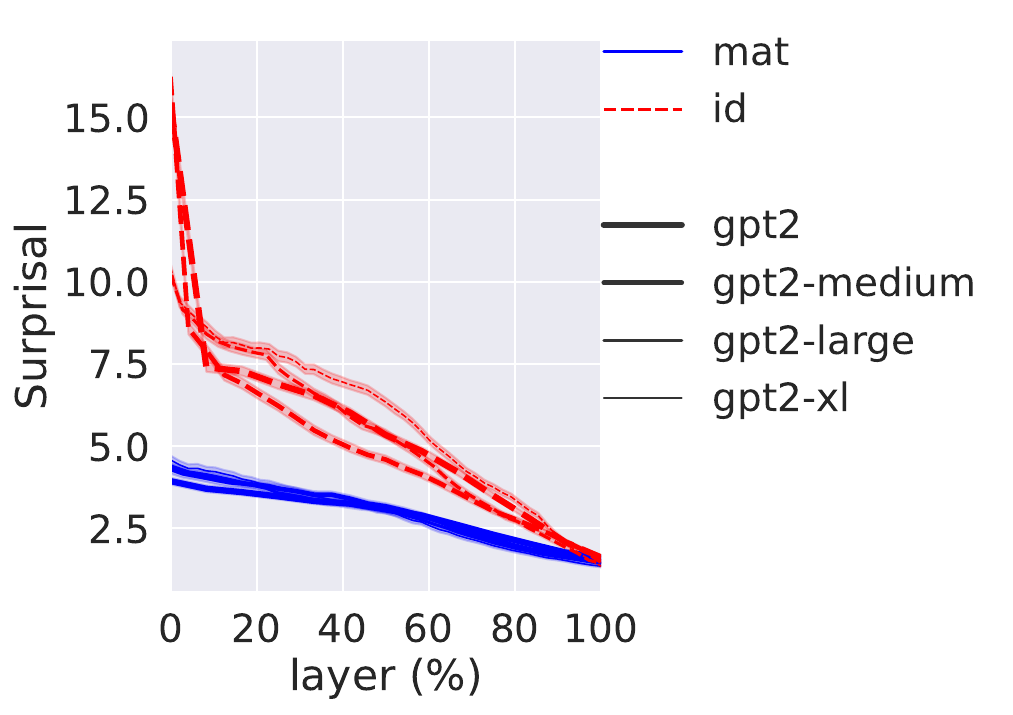}
\caption{Surprisal for $\matlL$ and $\idlL{}$ (\gpt{} next token prediction task, four model sizes). A 95\% confidence interval surrounds the lines.}
\label{fig:rob_surp}
\end{figure}
}

\quash{
\begin{figure}[t]
\centering
\includegraphics[scale=0.35]{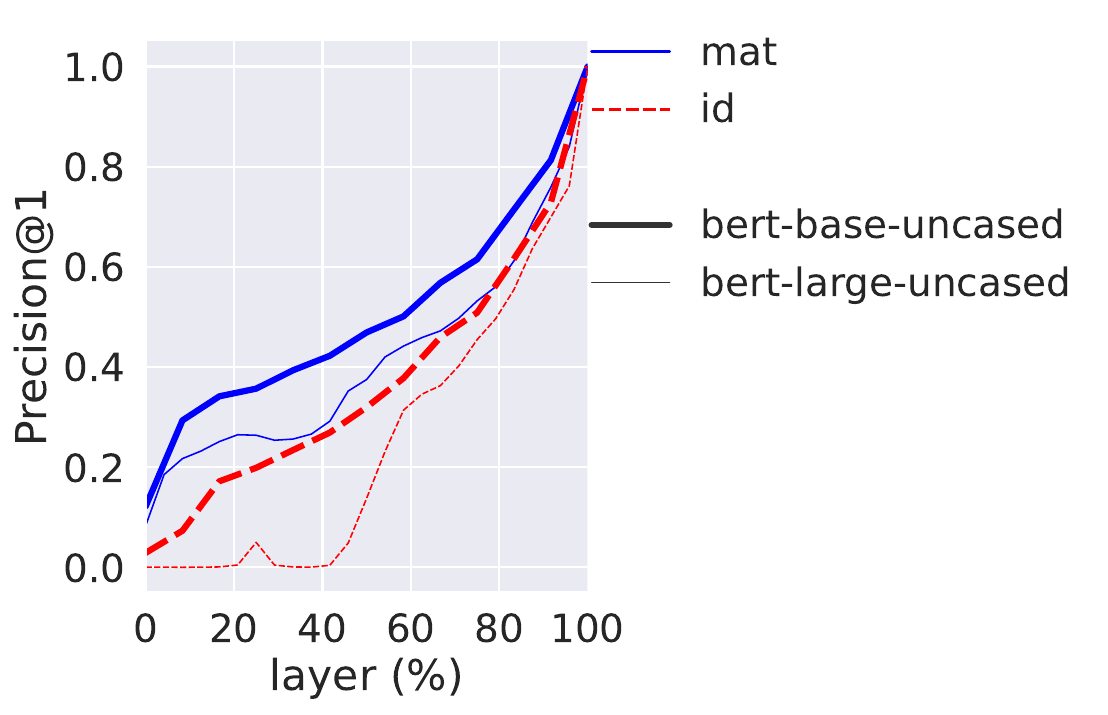}
\caption{Precision@$1$ for $\matlL{}$ and $\idlL{}$ (\bert{} masked token prediction task, two model sizes).}
\label{fig:rob_bert_pre}
\end{figure}

\begin{figure}[t]
\centering
\includegraphics[scale=0.35]{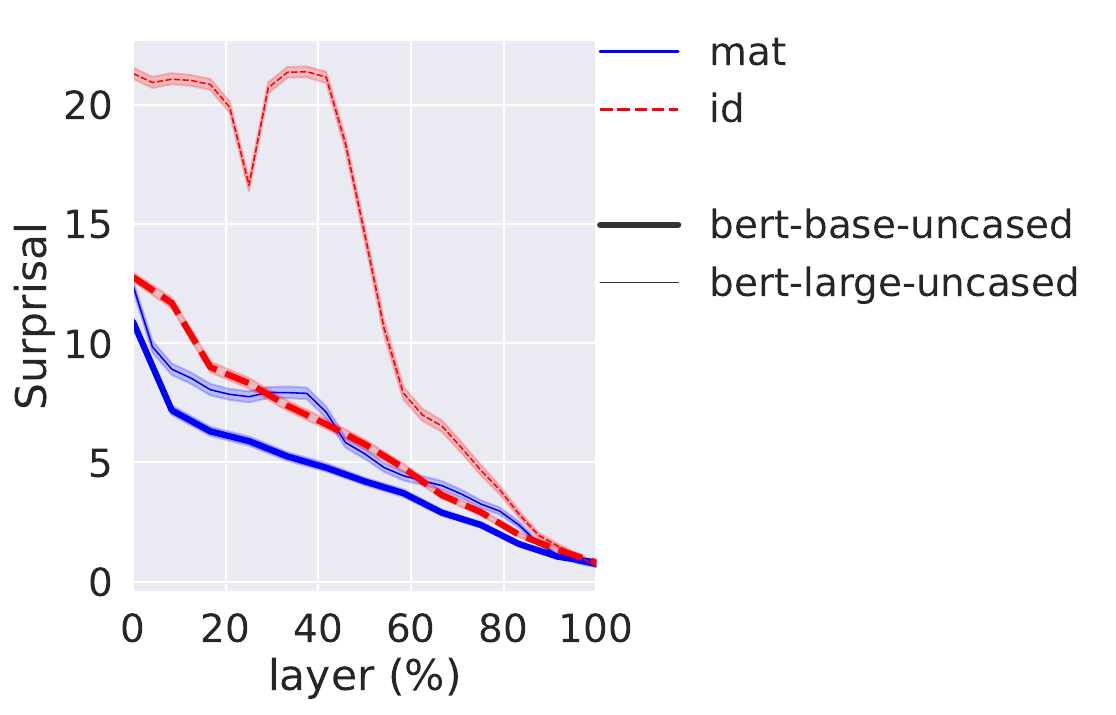}
\caption{Surprisal for $\matlL$ and $\idlL{}$ (\bert{} masked token prediction task, two model sizes). A 95\% confidence interval surrounds the lines.}
\label{fig:rob_bert_surp}
\end{figure}
}

\subsection{Robustness Across Data Distributions}
\label{subsec:robust_datasets}

We test whether the linear mappings learned from one data distribution are useful for predictions on other data distributions.
To this end, we use a second dataset of news article sentences, the 10K English 2020 news sentences corpus
from the Leipzig Corpora Collection \cite{goldhahn-etal-2012-building}, which we randomly divide into a training set $\mathcal{T}$ consisting of 9,000 examples and a validation set $\mathcal{V}$ consisting of 1,000 examples. For our experiments, we use the $24$-layer \gpt{} and \bert{} models.
First, we replicate previous experiments on the Leipzig dataset, obtaining results that are extremely similar; for example, the average (across layers) difference between the Precision@$1$ score of Wikipedia and Leipzig is $0.3\%$ for \gpt{} and $-1.4\%$ for \bert{}.
Next, we use Leipzig (resp. Wikipedia) samples to fit linear mappings $\matlL{}$ (as described in \S\ref{sec:layer_jump}), and then evaluate these mappings in the context of next-token prediction, on samples from Wikipedia (resp. Leipzig) (as in \S\ref{sec:prediction}). When swapping the original mappings with those trained on the other dataset, we observe a decrease of $0.1\%$ (resp. increase of $1.1\%$) relative to the original Precision@$1$ scores for \bert{} and a decrease of $5.5\%$ (resp. decrease of $8\%$) relative to the original Precision@$1$ scores for \gpt{}, on average across layers.
Overall, this shows that our method generalizes well to out-of-distribution samples. Moreover, our linear mappings capture general, rather than domain specific, features of the model's inference pass.
\section{Implication to Early Exiting}
\label{sec:applications}

The possibility of approximating
the final prediction already in the early layers has important implications for efficiency; applying our linear mapping instead of executing transformer blocks of quadratic time complexity, could save a substantial portion of the computation. In this section, we demonstrate this in the context of early exiting.

When 
using an early exit strategy \cite{schwartz-etal-2020-right, xin-etal-2020-deebert, schuster2022confident}, one aims at deciding dynamically at which layer to stop the computation and ``read'' the prediction from the hidden representation of that layer.
More precisely, under a confidence measure paradigm, one decides to stop the computation for a position $i$ at layer $\ell$ based on a confidence criterion, that is derived from casting the hidden representation $h_i^\ell$ as a final-layer representation and converting it to an output probability distribution. Specifically, following \citet{schuster2022confident}, a decision to exit is made if the difference between the highest and the second highest probabilities is bigger than $$ 0.9 \cdot \lambda + 0.1 \cdot {\rm exp} (-4 i / N),$$
where $N$ is the average length of the input until position $i_s$ for $s \in \mathcal{V}$, and $\lambda$ is a hyper-parameter.

\begin{figure}[t]
\setlength{\belowcaptionskip}{-10pt}
\centering
\includegraphics[width=\columnwidth]{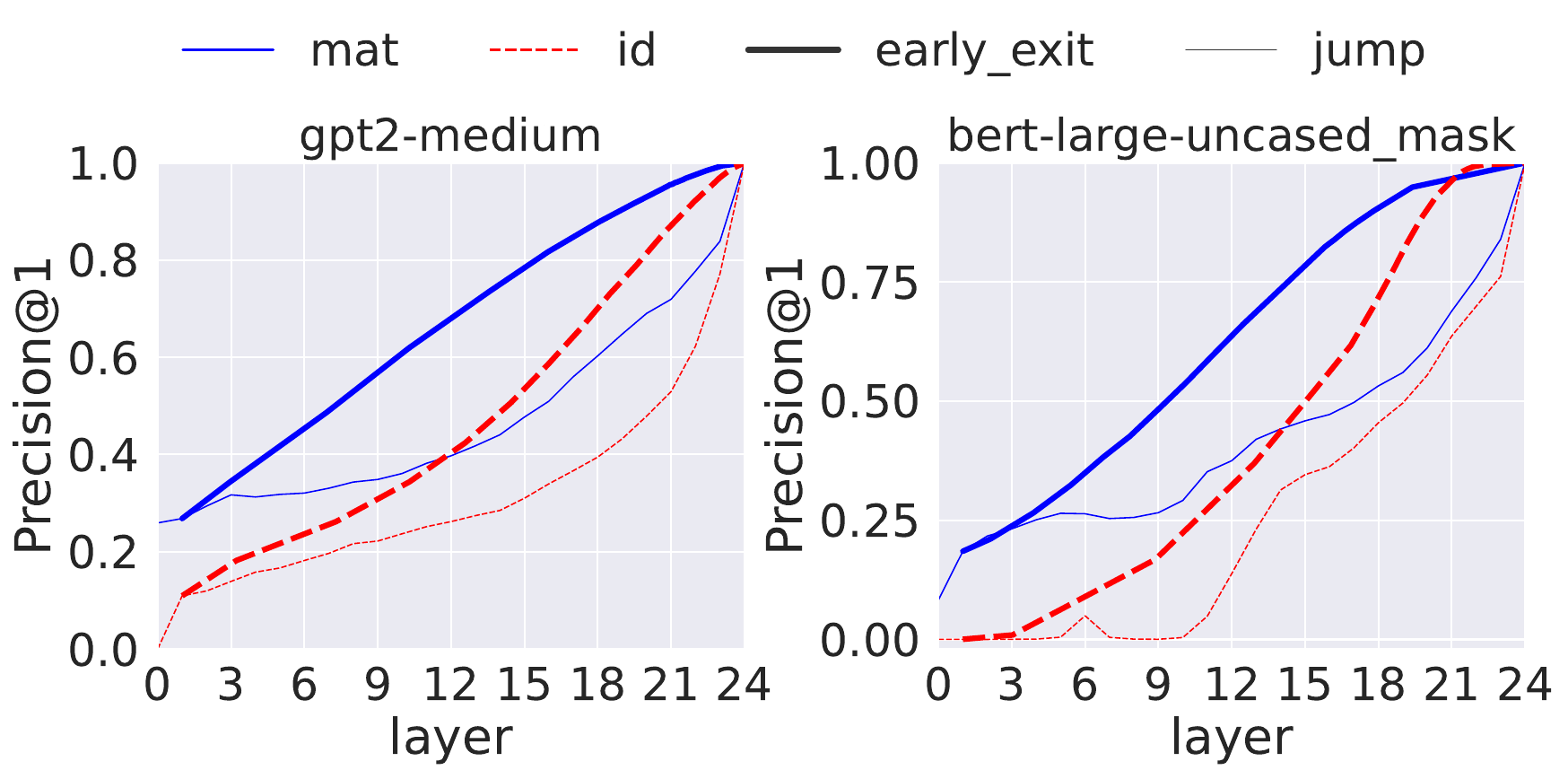}
\caption{Precision@$1$ with early exit and ``fixed exit'', applied to the $24$-layer \gpt{} for next token prediction (left) and the $24$-layer \bert{} for masked token prediction (right). Varying the confidence parameter $\lambda$, the $x$-coordinate is the average number of layers processed before an early exit decision is reached.}
\label{fig:ee_gpt2bert}
\end{figure}

\quash{
\begin{figure}[t]
\setlength{\belowcaptionskip}{-10pt}
\centering
\includegraphics[scale=0.35]{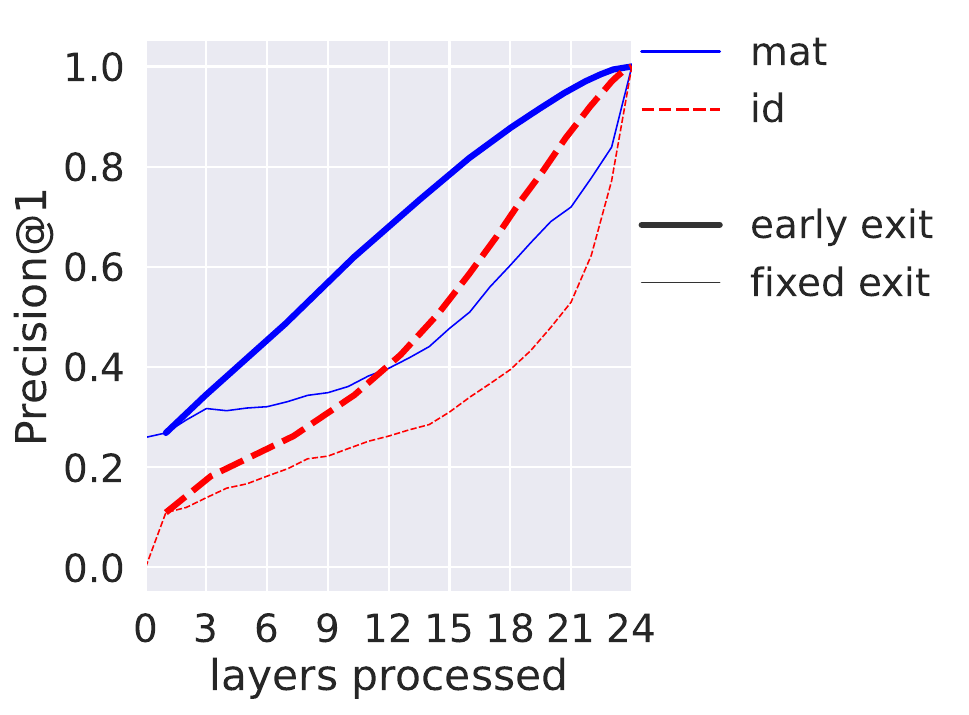}
\caption{Precision@$1$ for the various early exit methods, and previous ``fixed exit'' methods for comparison ($24$-layer \gpt{} next token prediction task). Varying the confidence parameter $\lambda$, the $x$-coordinate is the average number of layers processed before an early exit decision is reached.}
\label{fig:ee_pre1}
\end{figure}
}

\paragraph{Experiment.}
We assess the utility of our mapping $\matlL{}$ for early exit as a plug-and-play replacement for $\idlL{}$, through which intermediate representations are cast into final-layer representations.
We use \gpt{} for the next token prediction and \bert{} for masked token prediction (both with 24 layers).
We run each of the models over the validation set examples, while varying the confidence parameter $\lambda$ and using either $\idlL{}$ or $\matlL{}$ for casting intermediate representations.
Furthermore, we compare these early exit variants to the ``fixed exit'' strategy from \S\ref{sec:prediction}, where the computation is stopped after a pre-defined number of layers rather than relying on a dynamic decision.
We evaluate each variant in terms of both prediction's accuracy, using the Precision@$1$ metric (see \S\ref{sec:prediction}), and efficiency, measured as the average number of transformer layers processed during inference.

\paragraph{Results.}
Fig.~\ref{fig:ee_gpt2bert}
plots the average Precision@$1$ score against the average number of layers processed, for $24$-layer \gpt{} and $24$-layer \bert{}. For both models, under an early exit strategy our mapping \mat{} again provides a substantial improvement over \id{}.
For example, aiming at $95\%$ average precision, \mat{} saves $\sim3.3$ ($13.8$\%) layers in \gpt{} compared to only $\sim1.4$ ($5.9$\%) layers by \id{}, and $\sim4.8$ ($20$\%) layers in \bert{} versus $\sim3.5$ ($14.6$\%) layers by \id{}.
These results highlight the potential gains prominent early exit methods can obtain by using our method.
Notably, in both models and for each of the mapping methods, early exit obtains better results than fixed layer exit, as expected. 

\quash{
\begin{figure}[t]
\setlength{\belowcaptionskip}{-10pt}
\centering
\includegraphics[scale=0.35]{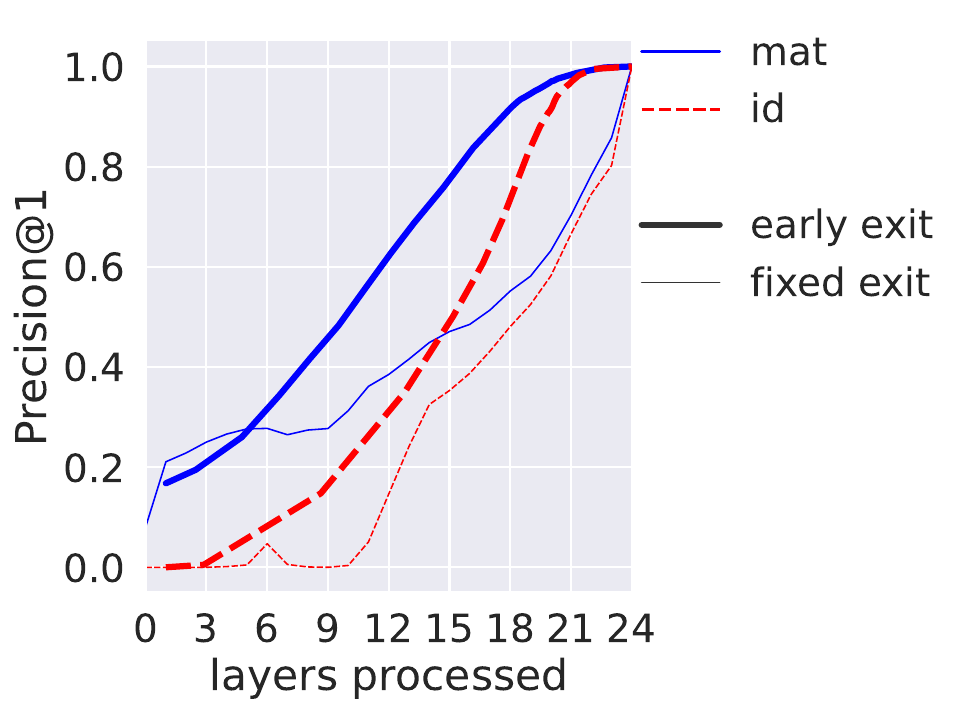}
\caption{Precision@$1$ for the various early exit methods, and previous ``fixed exit'' methods for comparison ($24$-layer \bert{} masked token prediction task). Varying the confidence parameter $\lambda$, the $x$-coordinate is the average number of layers processed before an early exit decision is reached.}
\label{fig:bertmask_ee_pre1}
\end{figure}
}
\section{Linear Shortcut Across Sub-Modules}
\label{sec:submodules}

In this section, we investigate whether discrepancies across layers result from specific sub-modules or are a general behaviour of all sub-modules in the network.  
This is done by extending our approach to test how well particular components in transformer blocks can be linearly approximated.

\paragraph{Method.}

Consider \gpt{} for definiteness, then:
$$ \texttt{b}_{\ell} = \texttt{b}_{\ell}^{\texttt{ffn}} \circ \texttt{b}_{\ell}^{\texttt{attn}}$$ 
\begin{equation}\label{eq:attn} \texttt{b}^{\texttt{attn}}_{\ell} (H) = \texttt{attn}_{\ell} (\texttt{ln1}_{\ell} (H)) + H,\end{equation} 
where $\texttt{attn}_{\ell}$ is
a MHSA
layer and \texttt{ln1} is a layer normalization (LN), and 
$$ \texttt{b}^{\texttt{ffn}}_{\ell} (H) = \texttt{ffn}_{\ell} (\texttt{ln2}_{\ell} (H)) + H,$$  
where $\texttt{ffn}_{\ell}$ is
an FFN
layer and $\texttt{ln2}$ is a LN.
\quash{
Given a block $\texttt{b}_\ell$ and one of its sub-modules $\texttt{ln1}_\ell, \ \texttt{attn}_\ell, \ \texttt{ln2}_\ell$, or $\texttt{ffn}_\ell$, we fit linear regression approximating the output of the sub-module given its input and then use it in order to define mappings, as we now describe.
}
Given a block $\texttt{b}_\ell$ and one of its sub-modules $\texttt{ln1}_\ell, \ \texttt{attn}_\ell, \ \texttt{ln2}_\ell$, or $\texttt{ffn}_\ell$, we fit linear regression approximating the output of the sub-module given its input, and then use it to define mappings $\matattnl{}$, $\matlnl{}$ and $\matffl{}$.
We provide the formal definitions of these mappings in App. \ref{sec:app_submodule_skip_description}.

\paragraph{Evaluation.}

We analyze the $24$-layered \gpt{}, and proceed completely analogously to \S\ref{subsec:next_token_prediction_task}, evaluating the Precision@$1$ and Surprisal metrics for the mappings $\matattnlL{}$, $\matfflL{}$ and $\matlnlL{}$.

\begin{figure}[t]
\setlength{\belowcaptionskip}{-0pt}
\centering
\includegraphics[width=\columnwidth]{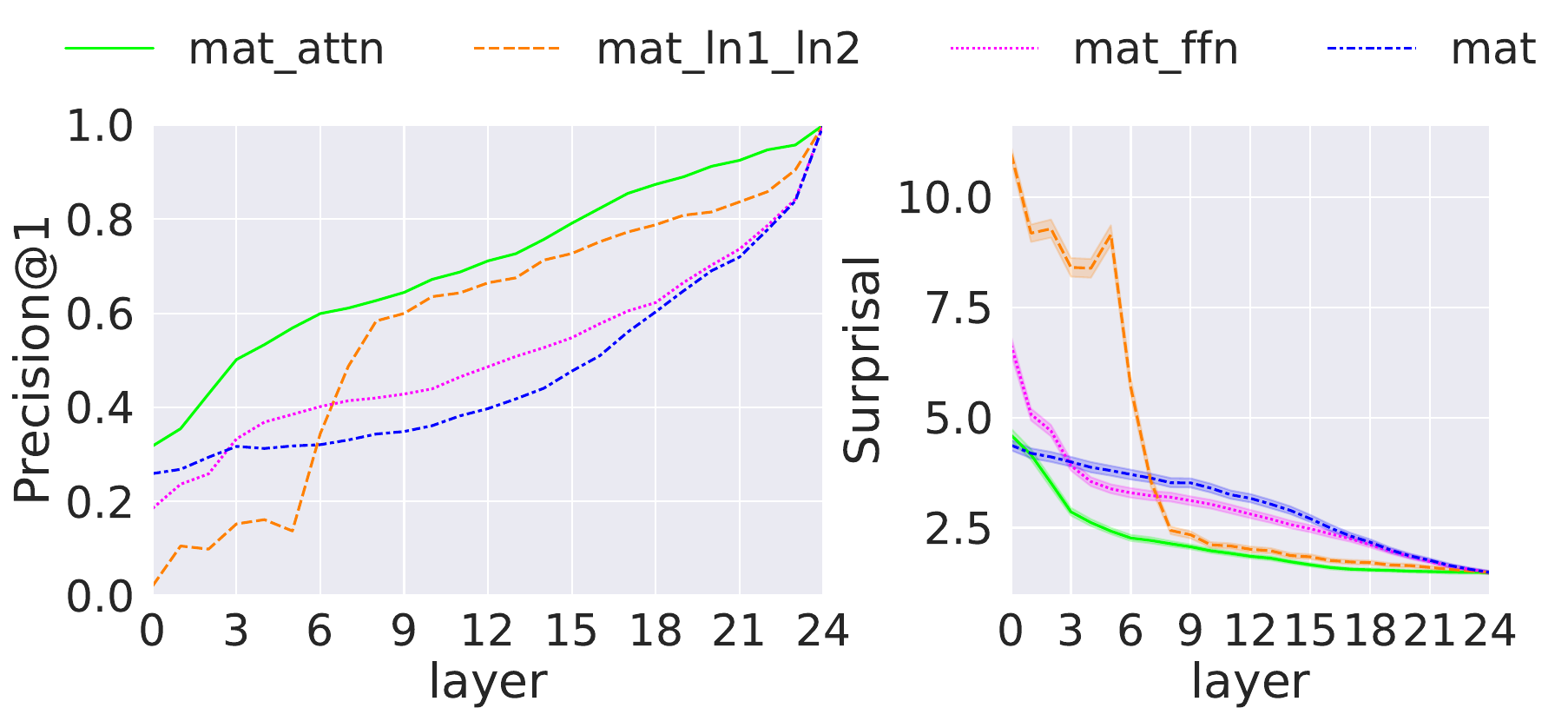}
\caption{Precision@$1$ and Surprisal for the various sub-module linear mappings, and $\matlL{}$ for comparison ($24$-layer \gpt{} next token prediction task). A 95\% confidence interval surrounds the Surprisal lines.}
\label{fig:parts_presurp}
\end{figure}

\quash{
\begin{figure}[t]
\centering
\includegraphics[scale=0.4]{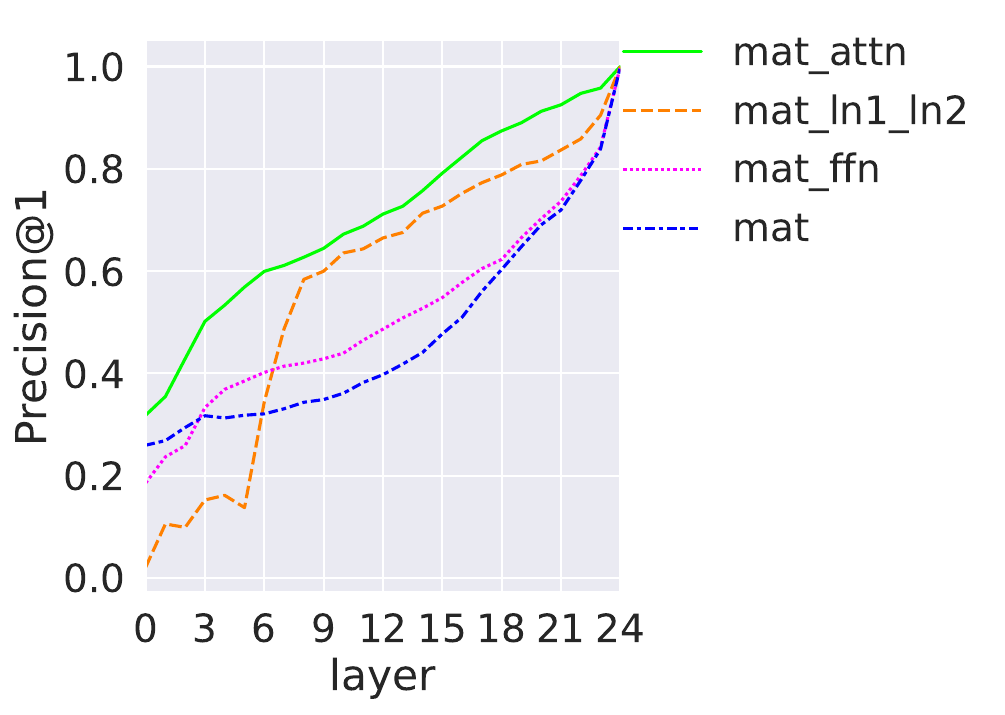}
\caption{Precision@$1$ for the various sub-module linear shortcut mappings, and the mapping $\matlL{}$ for comparison (\gpt{} next token prediction task).}
\label{fig:parts_pre1}
\end{figure}

\begin{figure}[t]
\centering
\includegraphics[scale=0.35]{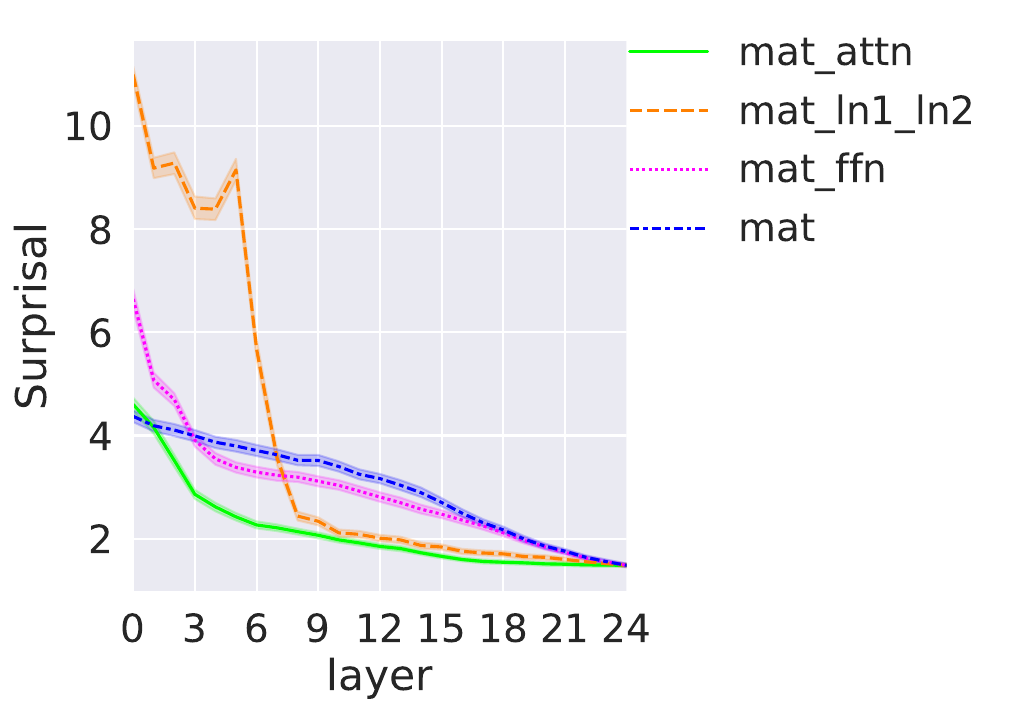}
\caption{Surprisal for the various sub-module linear shortcut mappings, and the mapping $\matlL{}$ for comparison (\gpt{} next token prediction task). A 95\% confidence interval surrounds the lines.}
\label{fig:parts_surp}
\end{figure}
}

\paragraph{Results.}
Fig.~\ref{fig:parts_presurp} shows the average Precision@$1$ and Surprisal scores per layer.
From a certain layer (\textasciitilde$7$), all sub-module mappings achieve better results than the full-block mapping $\matlL{}$. Thus, it is not just the cumulative effect of all the sub-modules in the transformer block that is amenable to linear approximation, but also individual sub-modules can be linearly approximated. 
Furthermore, the linear approximation of attention sub-modules is less harmful than that of the FFN or LN sub-modules. 
A possible reason is that the linear replacement of FFN or LN ``erodes'' the self-attention computation after a few layers. 
Moreover, the good performance of $\matattnlL{}$ suggests that contextualization often exhausts itself in early layers; speculatively, it is only in more delicate cases that the self-attention of late layers adds important information. Last, remark the sharp ascent of the scores for layer normalization in layers $5$-$8$, for which we do not currently see a particular reason. To conclude, we see that the possibility of linear approximation permeates
transformer components.

\section{Related Work}

There is a growing interest in utilizing intermediate representations of LMs for interpretability and efficiency.
For interpretability, one seeks to understand the prediction construction process of the model \cite{tenney-etal-2019-bert, voita-etal-2019-bottom, geva-etal-2022-transformer}, or the features stored in its hidden representations \cite{adi2017finegrained, conneau-etal-2018-cram,liu-etal-2019-linguistic}. Our work is different as it converts intermediate representations into a final-layer form, which is interpretable by design.

Previous works on early exiting cut the computation 
at a dynamically-decided earlier stage 
\cite{schwartz-etal-2020-right,xin-etal-2020-deebert,schuster2022confident,gera-etal-2023-benefits}, or a fixed network is utilized to parallelize inference \citep{leviathan2022fast, chen2023accelerating}. However, these methods propagate intermediate representations directly, which we show is substantially worse than our approach. Also, our method requires training of considerably fewer parameters than methods such as \citet{schuster-etal-2021-consistent}, which learn a different output softmax for each layer.  

Last, skipping transformer layers and analyzing the linearity properties of transformer components have been discussed in prior works \cite{Zhao2021of,mickus-etal-2022-dissect,wang-etal-2022-skipbert,lamparth2023analyzing}.
Specifically, a concurrent work by \citet{belrose2023eliciting} proposed to train affine transformations from hidden to final representations to increase model transparency. Our work is different in that we train \textit{linear} transformations \textit{across all layers}. Moreover, while \citet{belrose2023eliciting} use SGD for training while minimizing KL divergence, we use linear regression, which requires much less compute. It will be valuable to compare the accuracy of both methods.

\section{Conclusion and Future Work}

We present a simple and effective method for enhancing utilization of hidden representations in transformer-based LMs, that uses 
pre-fitted context-free and token-uniform linear mappings.
Through a series of experiments on different data sources, model architectures and scales, we show that our method consistently outperforms the prevalent practice of interpreting representations in the final-layer space of the model, yielding better approximations of succeeding representations and the predictions they induce, thus allowing a more faithful interpretation of the model's prediction-formation.
We demonstrate the practicality of our method for improving computation efficiency, saving a substantial amount of compute on top of prominent early exiting approaches. 
Also, by extending our method to sub-modules, 
we observe that replacing a part of the transformer inference by a non-contextual linear computation often results in a small deterioration of the prediction.
This opens new research directions for improving model efficiency,
including breaking the computation into parallel tasks.
\section{Limitations}

First, while it is possible to define many different mappings in-between layers, for example, affine or non-linear transformations, we focus on the ``simple'' case of linear transformations.
Our choice is motivated by the wide success of the simplest mapping (i.e. the identity baseline, of inspecting hidden representations in the same linear space), while we are asking if there is more linearity in transformer inference that can be exploited for interpretability.

Second, we find that there is more linear structure to parts of the transformer computation (both full layers and sub-modules) than could be explained solely by the residual connection. However, we do not elucidate a reason for that, leaving exploration of this interesting research question for future work.

Third, our experiments focus on post-hoc interpretability, that is, analyzing a trained model without changing its weights. Future work should also consider analyzing the utility of such linear mappings when those are integrated into the model training.

Last, in our experiments we use only data in English. Nonetheless, given the comprehensiveness of our experiments and the fact that our method does not rely on any language-specific features, we would expect our findings to hold in other languages as well.


\section{Bibliographical References}\label{sec:reference}
\bibliographystyle{lrec-coling2024-natbib}
\bibliography{anthology, custom}

\clearpage
\appendix
\appendix

\section{Descriptions of $\matattn{}$, $\matff{}$ and $\matln{}$}
\label{sec:app_submodule_skip_description}

Here we detail the definitions of the mappings $\matattnl{}$, $\matffl{}$ and $\matlnl{}$ utilized in \S\ref{sec:submodules}.

\paragraph{Description of $\matattnl{}$.}
For an input $s$, let $v^\ell_{i_s}$ be the vector at position $i_s$ in the output of $\texttt{attn}_\ell (\texttt{ln1}_\ell (H^{\ell - 1}))$. We denote by $A_\ell^{\texttt{attn}} \in \mathbb{R}^{d_h \times d_h}$ the matrix numerically minimizing 
$$ A \mapsto \sum_{s \in \mathcal{T}} || A \cdot \texttt{ln1}_\ell (h^{\ell-1}_{i_s}) - v^\ell_{i_s}||^2,$$
and define an attention sub-module replacement (Eq.~\ref{eq:attn}) by $$
\texttt{b}^{\overline{\texttt{attn}}}_\ell (h) \coloneqq A_{\ell}^{\texttt{attn}} \cdot \texttt{ln1}_\ell (h) + h. $$
We then define a mapping between two layers ${\ell \rightarrow \ell'}$ by:
$$ \matattnl{} (h) \coloneqq $$
$$ \texttt{b}^{\texttt{ffn}}_{\ell'} ( \texttt{b}^{\overline{\texttt{attn}}}_{\ell'} ( \ldots (\texttt{b}^{\texttt{ffn}}_{\ell+1} ( \texttt{b}^{\overline{\texttt{attn}}}_{\ell+1} (h)))\ldots)).$$ 
Namely, when applying each $\ell''$-th block, $\ell < \ell'' \leq \ell'$, we replace its attention sub-module $\texttt{attn}_{\ell''}$ by its linear approximation.
Importantly, unlike the original attention module, the approximation $\texttt{b}^{\overline{\texttt{attn}}}_\ell$ operates on each position independently, and therefore applying $\matattnl{}$ disables any contextualization between the layers $\ell$ and $\ell'$. Note that this is not the case for $\matffl{}$ and $\matlnl{}$, which retain the self-attention sub-modules and operate contextually.

\paragraph{Description of $\matffl{}$.}
Let $v^\ell_{i_s}$ be the vector at position $i_s$ in the output of $\texttt{ln2}_{\ell} (\texttt{b}_\ell^{\texttt{attn}} (H^{\ell - 1}))$, for a given input $s$. We denote by $A_\ell^{\texttt{ffn}} \in \mathbb{R}^{d_h \times d_h}$ the matrix numerically minimizing 
$$ A \mapsto \sum_{s \in \mathcal{T}} || A \cdot v^{\ell}_{i_s} - \texttt{ffn}_{\ell} (v^\ell_{i_s})||^2,$$
and define a replacement of the feed-forward sub-module $\texttt{b}_{\ell}^{\texttt{ffn}}$ by $$ \texttt{b}^{\overline{\texttt{ffn}}}_\ell (H) \coloneqq A_{\ell}^{\texttt{ffn}} \cdot \texttt{ln2}_\ell (H) + H.$$
We then define a mapping between two layers ${\ell \rightarrow \ell'}$ by:
$$ \matffl{} (H) \coloneqq $$
$$ \texttt{b}^{\overline{\texttt{ffn}}}_{\ell'} ( \texttt{b}^{\texttt{attn}}_{\ell'} ( \ldots (\texttt{b}^{\overline{\texttt{ffn}}}_{\ell+1} ( \texttt{b}^{\texttt{attn}}_{\ell+1} (H))\ldots)).$$

\paragraph{Description of $\matlnl{}$.}
Let $v^\ell_{i_s}$ be the vector at position $i_s$ in the output of $\texttt{b}^{\texttt{attn}}_{\ell} (H^{\ell - 1})$, for a given input $s$. We denote by $A_\ell^{\texttt{ln1}} \in \mathbb{R}^{d_h \times d_h}$ the matrix numerically minimizing 
$$ A \mapsto \sum_{s \in \mathcal{T}} || A \cdot h^{\ell}_{i_s} - \texttt{ln1}_{\ell} (h^\ell_{i_s})||^2$$ and we denote by $A_\ell^{\texttt{ln2}} \in \mathbb{R}^{d_h \times d_h}$ the matrix numerically minimizing $$ A \mapsto \sum_{s \in \mathcal{T}} || A \cdot v^{\ell}_{i_s} - \texttt{ln2}_{\ell} (v^\ell_{i_s})||^2.$$ We define a replacement of the block $\texttt{b}^{\texttt{attn}}_{\ell}$ by \begin{equation} \texttt{b}^{\overline{\texttt{ln1}}}_\ell (H) \coloneqq \texttt{attn}_{\ell} (A_{\ell}^{\texttt{ln1}} \cdot H) + H\end{equation} and we define a replacement of the block $\texttt{b}^{\texttt{ffn}}_{\ell}$ by \begin{equation} \texttt{b}^{\overline{\texttt{ln2}}}_\ell (H) \coloneqq \texttt{ffn}_{\ell} (A_{\ell}^{\texttt{ln2}} \cdot H) + H.\end{equation}
We then define a mapping between two layers ${\ell \rightarrow \ell'}$ by:
$$ \matlnl{} (H) \coloneqq $$
$$ \texttt{b}^{\overline{\texttt{ln2}}}_{\ell'} ( \texttt{b}^{\overline{\texttt{ln1}}}_{\ell'} ( \ldots (\texttt{b}^{\overline{\texttt{ln2}}}_{\ell+1} ( \texttt{b}^{\overline{\texttt{ln1}}}_{\ell+1} (H))\ldots)).$$

\end{document}